\documentclass[10pt,journal,compsoc]{IEEEtran}

\ifCLASSOPTIONcompsoc
  \usepackage[nocompress]{cite}
\else
  \usepackage{cite}
\fi
\ifCLASSINFOpdf
  \usepackage[pdftex]{graphicx}
\else
  \usepackage[dvips]{graphicx}
\fi

\usepackage{amsmath,amssymb} 
\usepackage{url}

\makeatletter
\let\MYcaption\@makecaption
\makeatother

\usepackage[font=footnotesize,labelfont=sf,textfont=sf]{subcaption}

\makeatletter
\let\@makecaption\MYcaption
\makeatother

\usepackage{multirow}
\usepackage{algorithm}
\usepackage{algpseudocode}
\usepackage[dvipsnames]{xcolor}
\usepackage{tikz}
\usetikzlibrary{calc}
\usepackage{soul}
\usepackage{array}
\usepackage{xspace}

\input{coloredAlgo.tex}

\newcommand{\norm}[1]{\left\lVert#1\right\rVert}

\DeclareMathOperator*{\aggregate}{aggregate}
\def\RANSAAC{\emph{RANSAAC}\xspace}

\DeclareMathOperator*{\argmin}{\arg\!\min}
\DeclareMathOperator*{\argmax}{\arg\!\max}

\providecommand{\norm}[1]{\left\lVert#1\right\rVert}

\definecolor{shadecolor}{RGB}{255,80,80}

\newcommand{\nada}[1]{}

\providecommand{\norm}[1]{\left\lVert#1\right\rVert}

\newcommand{\y}{\textbf{y}}

\makeatletter
\newcommand{\mypm}{\mathbin{\mathpalette\@mypm\relax}}
\newcommand{\@mypm}[2]{\ooalign{%
  \raisebox{.1\height}{$#1+$}\cr
  \smash{\raisebox{-.6\height}{$#1-$}}\cr}}
\makeatother

\begin{document}
\title{Accurate Motion Estimation through Random Sample Aggregated Consensus}

\author{Martin~Rais, 
		Gabriele Facciolo, 
        Enric~Meinhardt-Llopis,\\ 
        Jean-Michel~Morel, 
        Antoni~Buades, 
        and~Bartomeu~Coll
\IEEEcompsocitemizethanks{\IEEEcompsocthanksitem M. Rais, A. Buades and T. Coll are with the Departament de Matem\`{a}tiques i Inform\`{a}tica,  Universitat de les Illes Balears, Balearic Islands, Spain. E-mail: \{martin.rais, toni.buades, tomeu.coll\}@uib.es.\protect\\
\IEEEcompsocthanksitem G. Facciolo, E. Meinhardt-Llopis, and J.-M. Morel are with the CMLA, Ecole Normale Sup\'{e}rieure Paris-Saclay, 94230, France. E-mail:\{facciolo, enric.meinhardt, morel\}@cmla.ens-cachan.fr.
}
}

\markboth{IEEE TRANSACTIONS ON PATTERN ANALYSIS AND MACHINE INTELLIGENCE, ~Vol.~X, No.~X, August~2016}%
{Rais \MakeLowercase{\textit{et al.}}: Accurate Global Transformation Estimation through Random Sample Aggregated Consensus}

%

\IEEEtitleabstractindextext{%
\begin{abstract}
We reconsider the classic problem of estimating accurately a 2D transformation from point matches between images containing outliers. RANSAC discriminates outliers by randomly generating minimalistic sampled hypotheses and verifying their consensus over the input data. Its response is based on the single hypothesis that obtained the largest inlier support. In this article we show that the resulting accuracy can be improved by aggregating all generated hypotheses. This yields \RANSAAC, a framework that improves systematically over RANSAC and its state-of-the-art variants by statistically aggregating hypotheses. To this end, we introduce a simple strategy that allows to rapidly average 2D transformations, leading to an almost negligible extra computational cost. We give practical applications on projective transforms and homography+distortion models and demonstrate a significant performance gain in both cases.
\end{abstract}

}

\maketitle

\IEEEdisplaynontitleabstractindextext

%
\IEEEpeerreviewmaketitle

\IEEEraisesectionheading{\section{Introduction}\label{sec:introduction}}



\IEEEPARstart{S}{everal} applications in computer vision such as image alignment, panoramic mosaics, 3D reconstruction, motion tracking, object recognition, among others, rely on feature-based approaches. These approaches can be divided in three distinct steps. First, characteristic image features are detected on the input images. They can come either in the form of special image points \cite{Lowe2004} or as distinguished image regions \cite{matas2002robust}, and should meet some repeatability criteria to ensure they can be detected on images, regardless of their location, pose, illumination conditions, scale, etc. 
Second, a unique description is given to each detected feature. In the case of feature points, this is done by describing its surroundings \cite{Lowe2004, bay2008speeded, Mikolajczyk_2005}. For regions, their silhouettes and/or the contained texture information are employed \cite{belongie2002shape,forssen2007shape}. Finally, the descriptors of both images are matched together. Ideally, these matches correspond to samples of the underlying model to be estimated. Unluckily, the presence of outliers (i.e. matches that do not correspond to this model) complicates the estimation task. Moreover, the locations of matching points are frequently contaminated with noise commonly referred to as measurement noise \cite{raguram_exploiting_2009}.

Introduced by Fischler and Bolles \cite{Fischler_1981}, Random Sample Consensus (RANSAC) is an iterative method that deals with outliers, and is used to simultaneously solve the correspondence problem while estimating the implicit global transformation. 
By randomly generating hypotheses on the transform matching the points, it tries to achieve a maximum consensus in the input dataset in order to deduce the matches according to the transformation, known as inliers. Once the inliers are discriminated, they are used to estimate the parameters of the underlying transformation by regression. RANSAC achieves good accuracy when observations are noiseless, even with a significant number of outliers in the input data. 

Instead of using every sample in the dataset to perform the estimation as in traditional regression techniques, RANSAC tests in turn many random sets of sample pairs. Since picking an extra point decreases exponentially the probability of selecting an outlier-free sample \cite{chum_locally_2003}, RANSAC takes the minimum sample size (MSS) to determine a unique candidate transform, thus incrementing its chances of finding an all-inlier sample, i.e., a sample exclusively composed of inliers. This transform is assigned a score based on the cardinality of its consensus set. Finally the method returns the hypothesis that achieved the highest consensus. To refine the resulting transformation, a last-step minimization is performed using only its inliers.

\textbf{RANSAC weaknesses.}
The RANSAC method is able to give accurate results even when there are many outliers in the input data. However, it leaves some room for improvement. First, the probability of RANSAC obtaining a reasonable result increases with the number of iterations, however it may never reach the optimal solution. In addition, RANSAC results have a high degree of variability for the same input data, and this variability increases with the amount of input points and their measurement noise. Second, although robust to outliers, RANSAC is not particularly immune to measurement noise  on the input data \cite{raguram_exploiting_2009}. 
Furthermore, the final minimization step weighs uniformly the assumed inlier match, ignoring whether they are real inliers or how strongly they may be affected by noise.  Third, the maximum tolerable distance parameter $\delta_d$ should be sufficiently tight to obtain a precise transformation, however it must also be loose to find enough input samples \cite{Choi2009}. Because of such considerations, setting this parameter is a difficult task, even under low measurement noise. Fourth, even though the total amount of iterations can be set according to the theoretical bound proposed by the original authors \cite{Fischler_1981}, more iterations are typically required \cite{raguram_exploiting_2009}.
Finally and most importantly, \textbf{the accuracy of RANSAC is based on the single sample where the best model was found}. Although this may be accurate in some cases, it nevertheless ignores the other all-inlier models that may have been generated throughout the iterations, and which could be used to improve the resulting accuracy.

%
%

\textbf{Contributions.} To make up for these issues, we propose Random Sample Aggregated Consensus (\RANSAAC), a simple yet powerful method that combines the random sample consensus scheme with a statistical approach. By aggregating the random hypotheses weighted by their consensus set cardinalities, the proposed approach improves systematically on RANSAC. We give practical implementations of this idea on 2D parametric transformation models, by proposing a simple strategy that allows to rapidly aggregate 2D parametric transformations using negligible computational resources. 

Its main benefits over traditional RANSAC are fourfold. 
First, results are both more accurate and with less variability (in terms of standard deviation of the error). The improvement in accuracy over the traditional approach is on average by a factor of two to three, and is even more important with higher noise, more inliers available, and higher outlier ratios. Moreover, it improves systematically over other state-of-the-art RANSAC extensions.
Second, as with the original RANSAC method, the accuracy is dramatically improved by adding a local optimization step, which in fact seems suited for our approach because it avoids discarding the potentially useful intermediate models generated during its execution.
Third, by including this step, the theoretical adaptative stopping criterion proposed in \cite{Fischler_1981}  
becomes more realistic and could be used to effectively stop the iterations without affecting the final accuracy.
Finally, using the proposed 2D transformation aggregation method adds an almost negligible extra computational cost to the RANSAC algorithm.


The rest of this article is organized as follows. We begin by giving a review on RANSAC and its variants in section \ref{sec:RANSACandItsVariants}. In section \ref{sec:aransac} we detail the proposed algorithm. We thoroughly evaluate it in section \ref{sec:evaluation} to finally conclude and give some remarks regarding the future work in section \ref{sec:conclusions}.

\section{RANSAC and its variants}
\label{sec:RANSACandItsVariants}
\subsection{The RANSAC method}
Given an unknown transformation $\phi$ with parameters $\theta$ and a dataset of pairs $X_i,Y_i$  consisting of $i=1, \dots, N$ samples, RANSAC computes
\begin{equation}
\displaystyle \hat{\theta} = \argmax_\theta \sum_{i=1}^N \rho\Big(dist\big(\phi_{\theta}(X_i),Y_i\big)\Big),
\label{eq:RANSAC}
\end{equation}
where $dist$ is usually the squared $L_2$ distance 
and the cost function $\rho$ is defined as
\begin{equation}
\rho(e) = \left\{
\begin{array}{c c}
1 & \ \ \text{if } e \leq \delta_d,\\
0 & \ \ \text{otherwise.}
\end{array}
\right.
\end{equation}
The parameter $\delta_d$ is a critical parameter of the algorithm and will be discussed below. 
Finally, the optional last-step minimization refines the previous result by computing
\begin{equation}
\hat{\theta} = \argmin_\theta \sum_{i=1}^M dist\big(\phi_{\theta}(\tilde{X}_i),\tilde{Y}_i\big),
\label{eq:RANSAC+M}
\end{equation}
where the $M$ inlier matches $(\tilde{X}_i,\tilde{Y}_i)$ are defined by
\begin{equation}
\{(\tilde{X}_i,\tilde{Y}_i)~|~ \tilde{X}_i \in X, \tilde{Y}_i \in Y, \rho(dist\big(\phi_{\hat{\theta}}(\tilde{X}_i),\tilde{Y}_i\big)) = 1 \}.
\end{equation} 

\subsubsection{RANSAC iterations and model error}
\label{sec:adaptiveTermination}
The authors of RANSAC \cite{Fischler_1981} show that by uniformly drawing $k$ samples, with
\begin{equation}
k \geq \frac{\log(1-\eta_0)}{\log(1 - \epsilon^m)},
\label{eq:amountOfIterations}
\end{equation}
where $m$ is the minimum sample size (MSS), one can ensure with confidence $\eta_0$ that at least one outlier-free sample is obtained from a dataset having inlier rate $\epsilon$. Therefore, the total number of iterations could be set according to Eq.~\eqref{eq:amountOfIterations}. However, since the inlier rate $\epsilon$ is in general not known beforehand, 
this value is updated every time a better hypothesis is found, by using Eq.~\eqref{eq:amountOfIterations} and taking $\epsilon$ as the current inlier ratio.

However, as noted by Chum \emph{et al.} \cite{chum_locally_2003}, the number of iterations of Eq.~\eqref{eq:amountOfIterations} is overly optimistic, so that an outlier-free sample does not guarantee an accurate estimation either. This is due not only to the noise on the position of the input points, but also to the underlying noise on the model hypotheses themselves \cite{Tordoff_2002,raguram_usac_2013}. This other noise is caused by several factors such as limited numerical precision, poor conditioning of the model estimation method, or by other sometimes non-avoidable reasons such as the rounding of pixel intensities or point coordinates. In practice, the number of samples required in RANSAC should be set typically to two or three times the theoretical number \cite{raguram_exploiting_2009}.

 \subsubsection{Distance parameter}
 \label{sec:distanceParam}
The parameter $\delta_d$ depends on the measurement noise of the input samples. Assume that these are 2D points with Gaussian noise of zero mean and standard deviation $\sigma$. Then the direct transfer error, computed as 
\begin{equation}
dist(\phi_\theta(X_i), Y_i) = \norm{Y_i- \phi_{\theta}(X_i)}^2,
\end{equation}
i.e. the distance between the projected point on the first image onto the second image and its match, is the sum of squared Gaussian variables which follows a $\chi^2_m$ distribution with $m\!=\!2$ degrees of freedom. The probability that this error is lower than a given value $k$ can be computed from  the cumulative $\chi^2_m$ distribution $F_m$.
Capturing a fraction $\alpha$ of the inliers is ensured by setting
$\delta_d = F_m^{-1}(\alpha)\sigma^2$,
which means that a true inlier will be incorrectly rejected $1-\alpha$ percent of the time. Then if the error of the estimated transformation $\phi$ for a particular set of matches ($X, Y$) is measured as the direct transfer error, 
this implies setting $\delta_d = 5.99 \sigma^2$ for $\alpha=0.95$ \cite{Hartley_2003}. However, if the error is given by the symmetric transfer error, namely
\begin{equation}
dist(\phi_\theta(X_i), Y_i) = \norm{Y_i, \phi_{\theta}(X_i)}^2 + \norm{X_i, \phi^{-1}_{\theta}(Y_i)}^2,
\label{eq:symmetricTransferError}
\end{equation}
then the $\chi^2_m$ distribution has four degrees of freedom (i.e. $m=4$). Therefore, one must set $\delta_d = 9.4877 \sigma^2$ for $\alpha=0.95$ or $\delta_d = 13.2767\sigma^2$ for $\alpha=0.99$. However, since in practice, the standard deviation of the noise is not available, this parameter is usually set experimentally. Throughout this article, this parameter was fixed to $\alpha=0.99$ and the error was assumed to be measured as in Eq.~\eqref{eq:symmetricTransferError}.

\subsubsection{Final refinement step} The original RANSAC method reduces the effect of measurement noise by re-estimating the transform using a last-step minimization on the resulting inliers. Depending on the underlying model, both linear and non-linear minimization methods exist \cite{Hartley_2003,botterill2011refining}, with the latter one achieving better results with a higher computational cost. Although this final stage does in general improve RANSAC results, it gives equal importance to each match for the minimization. Then if RANSAC is not able to completely discriminate the outliers within the input data, this introduces a bias in the computation often resulting in a performance drop.

Throughout this article, we will refer to RANSAC as the method described by Eq.~\eqref{eq:RANSAC} without performing the last refinement step, and RANSAC+M or more directly RANSAC with last-step minimization to the full approach following Eq.~\eqref{eq:RANSAC+M}.

\subsection{RANSAC Alternatives or Improvements}
\label{sec:state-of-the-art}
One of the aims of robust statistics is to reduce the impact of outliers. In computer vision, a robust estimator must be able to find the correct model despite of the noisy measurements and the high percentage of outliers.

In the context of regression analysis, the Least-Median-of-Squares (LMedS) and the Least Trimmed Squares (LTS) are well-known classical methods and still widely used \cite{Rousseeuw_84}. Both approaches resemble RANSAC by generating hypotheses from  minimalistic samples. However instead of discarding the outliers, both methods sort the input points based on their residual error. The LMedS then returns the hypothesis with the lowest residual of the median point. The LTS method returns instead the model for which the sum of a fixed percentage of best ranked input data is minimized. Although the LMeds method does not require to specify the $\delta_d$ parameter of RANSAC, it implicitly assumes at least $50\%$ of inliers in the dataset. This method was successfully used to estimate the epipolar geometry between two images \cite{Zhang_95}.

Since its introduction and due to the massive proliferation of panorama generation \cite{Brown2006}, 3D estimation \cite{Hartley_2003} and invariant feature points \cite{Lowe2004,IVES_2014}, several modifications of RANSAC were proposed \cite{raguram_comparative_2008,raguram_usac_2013,Choi_09}, mainly addressing three major aspects: computational cost, accuracy, and robustness against degenerate cases, i.e., cases where it is not possible to recognize an incorrect hypothesis by simply scoring the drawn minimal sample. Interestingly, these techniques can be grouped based on the strategy they adopt.

One of such strategies is to \emph{modify the random sampling process} of RANSAC to generate better hypotheses, improving both \emph{speed and accuracy}. In this group, NAPSAC \cite{Myatt2002NAPSACHN} segments the sample data by assuming that inliers tend to be spatially closer to one another than outliers, thus picking points laying within a hypersphere of a specific radius. The PROSAC algorithm \cite{chum_matching_2005} exploits the ordering structure of the set of tentative correspondences and bases its sampling on the similarity computed on local descriptors. GroupSAC \cite{NI_2009} combines both strategies by segmenting the input data based on feature characteristics. 
Suter and his collaborators show that taking random non-minimal subsets of samples considerably reduces the bias caused by taking minimal size samples over noisy input data \cite{Pham2014, Tennakoon_2016}. Another approach \cite{Chin_2012} is to perform an initial, not so extense round of random samples together with their hypotheses to build some prior knowledge, followed by guiding the subsequent sampling using the residuals of these hypotheses, i.e. the residual sorting information.

Another strategy is to perform a simple test by partially evaluating the dataset to \emph{discard hypotheses before computing the score}, thus reducing the overall \emph{computational cost}. The $T_{d,d}$ test \cite{matas_randomized_2004} checks whether $d$ randomly drawn points are inliers to the hypothesis, discarding it otherwise. In \cite{chum_optimal_2008}, the authors propose to use a sequential probability ratio test (SPRT) to discard ``bad'' hypotheses, based on Wald's theory of sequential testing \cite{wald1947sequential}. Finally, instead of evaluating one model at a time, Preemptive RANSAC \cite{nister_preemptive_2005} filters hypotheses by scoring several of them in parallel over the same subset of samples, permitting real-time applications.

To \emph{improve on accuracy}, instead of computing the cardinality of the set of sample points having smaller residuals than a specified threshold, several methods \emph{modify how samples are scored} by weighting inliers based on their residual error. The MSAC method \cite{Torr_1998}, a simple redescending M-estimator \cite{Huber81a}, takes the maximum between the error and the threshold to score each datum to finally minimize over their sum. Conversely, MLESAC \cite{torr_mlesac_2000} maximizes the log likelihood of the given data by assuming the inlier error has Gaussian distribution while the outlier error follows an uniform law.

One strategy used to improve both \emph{accuracy and speed} is to perform a \emph{local optimization} step after a ``good'' enough model has been found. The locally optimized RANSAC (LO-RANSAC) \cite{chum_locally_2003}, as well as its variants \cite{lebeda_fixing_2012,raguram_exploiting_2009}, performs a fixed amount of iterations by taking non-minimal samples of the inliers of each newly found best hypothesis, in a so-called \emph{inner}-RANSAC procedure. For each model generated by taking a non-minimal sample, a larger distance threshold is used to compute its inliers and an iterative procedure is performed to refine this model by progressively shrinking the threshold.
The recently published Optimal RANSAC method by Hast \emph{et al.} \cite{Has13a} also exploits this technique, however they add small modifications such as dataset pruning to remove outliers among iterations along with adaptive termination of the method when the same inlier set is found again after the \emph{inner}-RANSAC procedure.

To increase the \emph{robustness} of RANSAC against \emph{degenerate} data in an epipolar geometry estimation context, DEGENSAC \cite{Chum_2005_dominant_plane} performs a test aimed at identifying hypotheses where five or more correspondences in the minimal sample are related by a homography, to avoid bad solutions on images containing a dominant plane in the scene. In cases where most of the data samples are degenerate and do not provide a unique solution, QDEGSAC \cite{Frahm_2006} performs several RANSAC executions, iteratively, by adding constraints to the inliers of the previous results.

Based on these approaches, Raguram \emph{et al.} \cite{raguram_usac_2013} studied several RANSAC extensions and proposed an algorithm that intelligently combined them into a single modular method called Universal RANSAC (USAC). This method obtains state-of-the-art results using few computational resources. It does so by combining ideas from LO-RANSAC, DEGENSAC, the SPRT test and PROSAC. 

Another recent trend is to approach the robust \emph{estimation} problem in an \emph{inverse fashion} \cite{Toldo2008, Magri_2014, Magri_2015, Tepper_2014}. The idea is to use the residuals for each data point obtained from several random models to generate the \emph{preference matrix}, which relates input points with the generated hypotheses. This matrix is finally clustered to distinguish between inliers and outliers.
The methods in the literature mainly differ in how to sample, define, and cluster this matrix. While J-Linkage \cite{Toldo2008} uses a binary method to define the relationship between each data point and each hypothesis in the matrix, T-Linkage \cite{Magri_2014} uses a continuous thresholded value, conceptually similar to the difference between RANSAC and MSAC scores.  To perform the clustering, both methods rely on an agglomerative clustering technique: 
each iteration of the algorithm merges the two clusters with the smallest distance. J-Linkage uses the Jaccard distance metric while T-Linkage relies on the Tanimoto distance. 

The approach of Tepper and Sapiro \cite{Tepper_2014} performs bi-clustering on the preference matrix by using non-negative matrix factorization (NMF). 
A similar approach is used in \cite{Magri_2015}, where they combine Robust PCA with Symmetric NMF to cluster the preference matrix. It should be noted that, opposite to RANSAC where a single model is estimated, these approaches target multimodel estimation and therefore are not the focus of the present work.

Finally, to address the selection of the $\delta_d$ parameter, ORSA \cite{moisan_2012} proposes an \emph{a-contrario} criterion to avoid setting hard thresholds for inlier/outlier discrimination. On the other hand, Raguram \cite{raguram_exploiting_2009}  estimates the covariance of the underlying transform and propagates this error due to the noise by modifying the regions for which each data point is considered an inlier. 

\section{Random Sample Aggregated Consensus}
\label{sec:aransac}
As explained above, the RANSAC algorithm as well as its extensions discard the current best hypotheses when a new hypothesis with larger consensus is found.  
However, these hypotheses 
could be useful to obtain a more precise estimation of the underlying model. In fact, if hypotheses generated by all-inlier noisy samples are similar enough, i.e., their variance is bounded, then it makes sense to aggregate them to improve the accuracy of the method. 
The idea behind the proposed \RANSAAC approach is to take advantage of every generated hypothesis, using its score as a confidence measure allowing not only to discard outliers but also to give more importance to better models. In this section we first discuss the aggregation strategy employed, followed by presenting the local optimization improvement.

\subsection{Aggregation of 2D parametric transformations}
\label{sec:aggregation}
In the context of RANSAC, we shall first illustrate this idea through a simple but classic example. Assume again that two images are related by a homography $H$, each with some detected feature points and a certain amount of matching pairs between both images.  Let us consider a single feature $\mathbf{x}_0$ from the first image and its corresponding point $\mathbf{y}_0 = H(\mathbf{x}_0)$ on the second image (the match ($\mathbf{x}_0, \mathbf{y}_0$) does not necessarily need to be a detected match). At iteration $k$ RANSAC randomly picks matches, generates a hypothesized homography $H_k$  and computes its score using the whole dataset. This hypothesis in fact yields a tentative match  $\mathbf{\hat{y}_0} = H_k(\mathbf{x}_0)$. If the matches randomly selected were all inliers, then, with high probability, $\mathbf{\hat{y}_0}$ will be close to the real matching point $\mathbf{y_0}$ on the second image. Thus, after several lucky all-inlier iterations, many estimates close to $\mathbf{y_0}$ are actually found.

All of these estimates are affected by two noise sources: the underlying noise of the model hypothesis as explained before, and the measurement noise of the input points. Provided both of them have zero mean and a symmetric distribution function, aggregating these estimates should yield a value closer to the true point $\mathbf{y_0}$. However, since there may be outliers in the input samples, it is better to assign a weight for each generated hypothesis reflecting its reliability. Luckily, RANSAC already computes this measure, which is its so-called score. Therefore, by weighting each of the resulting estimated points using the RANSAC score and aggregating them by, for example, taking their weighted mean, the aggregated point will, with a high probability, be closer to its true value $\mathbf{y_0}$.

Therefore, in order to aggregate the transformations, the proposed approach requires the possibility to perform operations, such as averaging, directly on the models. 
Often, the set of these models has typically the structure of a Riemannian manifold \cite{Michor2012, robertron}. So they could be directly aggregated inside this manifold, by understanding its geodesics \cite{li2009projective}. However, for the case of 2D transformations mapping points in one image to points on another, instead of aggregating on this manifold, a simple strategy based on pre-selecting points could be used, virtually adding no cost to the overall computation. 


Let $\{X_i,Y_i\}_{i=1,\dots N}$ be a set of matching points where $X_i \in \mathcal{C}_1$, $Y_i \in \mathcal{C}_2$, both $\mathcal{C}_1, \mathcal{C}_2 \subset \mathbb{R}^2$ and $\{x_j\}_{j \in 1, \dots, n} \in \mathcal{C}_1$ with $n << N$ predefined points on the first image. For each iteration $k$ with hypothesis $\mathcal{H}_k = \phi_{\theta_k}$ \RANSAAC computes the projection of the predefined points
\begin{equation}
\hat{y}_j^k = \phi_{\theta_k}(x_j) \ \ \text{for } 1 \leq k \leq n
\end{equation}
and the RANSAC score
\begin{equation}
w_k = \sum_i^N \rho\Big(dist\big(\phi_{\theta_k}(X_i),Y_i\big)\Big),
\end{equation}
where $dist$ is again the squared $L_2$ distance 
and $\rho$ is the cost function. After $K$ iterations, for each predefined point $x_j$, \RANSAAC aggregates the different estimations $\hat{y}_j^k$ using the weights $w_k$, resulting in trustworthy matches $(x_j, \hat{y}_j)$
\begin{equation}
\hat{y}_j = \aggregate_{i=1, \dots, K}(\hat{y}_j^i, w_i)
\end{equation}
This procedure, in fact, could be regarded as a way to generate denoised inlier matches, since $x_j$ could be any point on the first image, while $\hat{y}_j$ will be placed, with a high probability, close to the true matching point $y_j$.

\subsubsection{Predefined Source Points}
\label{sec:predefinedSrcPoints}
\begin{figure}
  \centering
\begin{subfigure}{0.20\textwidth}
  \centering
  \includegraphics[width=\linewidth]{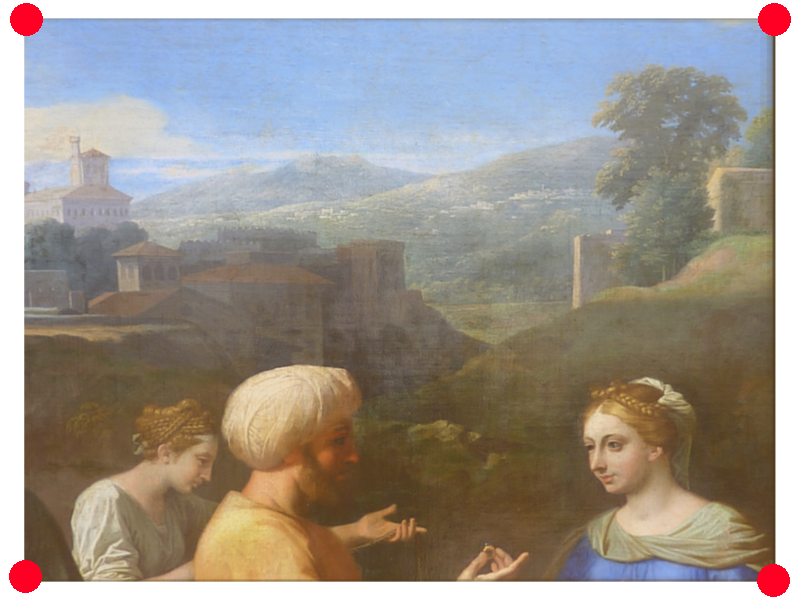}
\end{subfigure}%
\begin{subfigure}{0.20\textwidth}
  \centering
  \includegraphics[width=\textwidth]{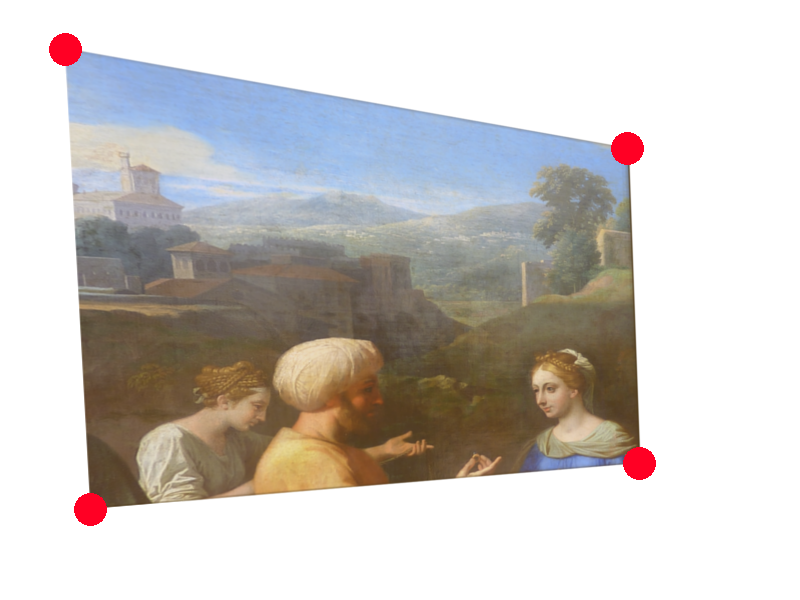}
\vspace{1mm}
\end{subfigure}
\begin{subfigure}{0.20\textwidth}
  \centering
  \includegraphics[width=\linewidth]{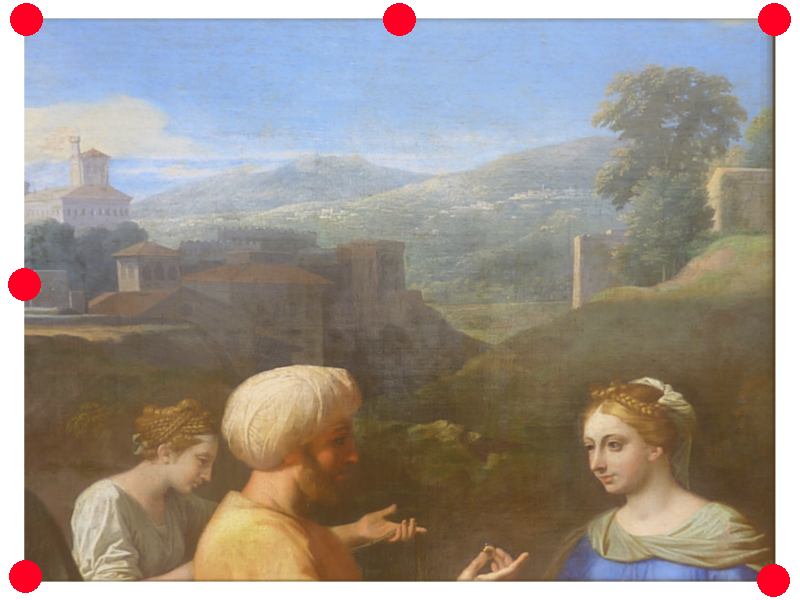}
\end{subfigure}%
\begin{subfigure}{0.20\textwidth}
  \centering
  \includegraphics[width=\textwidth]{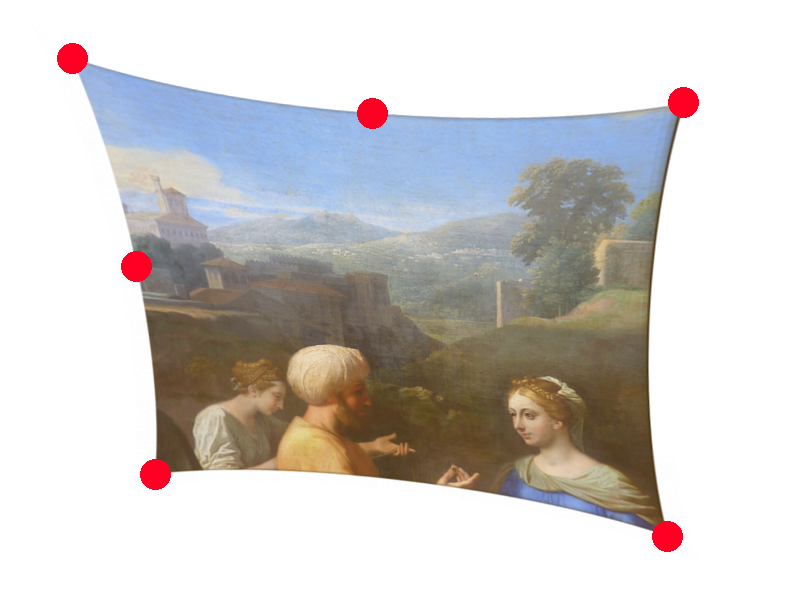}
\end{subfigure}
\caption{Suggested predefined source points (in red). \textbf{Top}: for homography estimation. \textbf{Bottom}: for homography + distortion.}
  \label{fig:examplePoints}
\end{figure}

The source points that are a parameter of the method depend on the global transform being estimated. The cardinality of this basis should be at least equal to the MSS. We select points which can uniquely determine the desired transform and minimize the noise caused by model generation, as depicted in Fig.~\ref{fig:examplePoints}. This strategy could be applied to any model that can be correctly estimated from points sampled on the image extents. For a homography, for example, the four points on the corners of the image should be selected. For a transformation involving a homography and two radial distortion parameters such as the one recently proposed by Kukelova et al. \cite{Kukelova_2015}, an additional point on the center of the top and\!/\!or the left image border must be added.

\subsubsection{Aggregation Strategies}
\label{subsec:aggregation}
\begin{figure}
\centering
\includegraphics[width=0.5\textwidth]{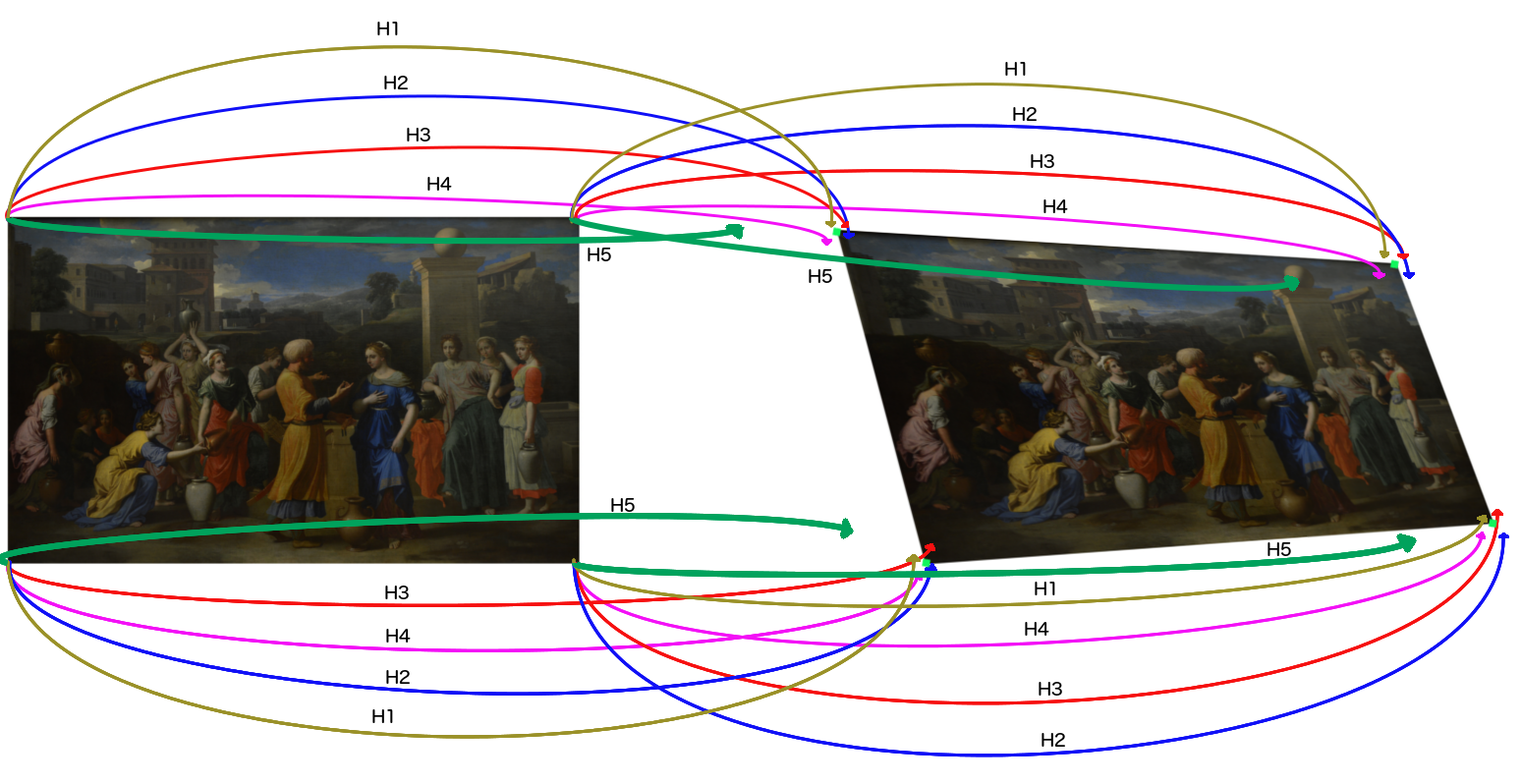}
\caption{Two images related by a homography. Each of the points on the corner of the first image are evaluated with 5 hypotheses. These destination points are aggregated to compute the final transform. }
\label{fig:example5Hypothesis}
\end{figure}

For the case of 2D parametric transformations, instead of simply returning the best hypothesis, \RANSAAC gathers, for each point of the predefined basis, a set of estimates of its location on the second image. These points form a  weighted cluster, each one being associated the score of the hypothesis from which it was derived. All of these estimates  can then be aggregated to compute a more accurate location for each point of the basis. This yields a minimum sample that permits to recover the transform. In the example shown in Fig.~\ref{fig:example5Hypothesis}, five possible hypotheses are shown. Since the fifth hypothesis $H_5$ (green arrow) obtained a lower score, it has less influence on the resulting aggregation.

Two aggregation techniques were considered, namely the weighted mean and the weighted geometric median  \cite{drezner_facility_2001}. To aggregate the points $x_i$ of the set $\mathcal{C}$, with weights $w_i$, with $i \in \{1, \dots, |\mathcal{C}|\}$, the weighted mean computes
\begin{equation}
wmean(\mathcal{C}) = \left(\displaystyle \sum_{i=1}^{|\mathcal{C}|} w_i^p x_i\right) \bigg/ \left(\displaystyle \sum_{i=1}^{|\mathcal{C}|} w_i^{p} \right),
\end{equation}
where $p$ is a parameter of the aggregation procedure.

The weighted geometric median of a set is defined as the point $y$ with the minimum sum of distances to the others
\begin{equation}
wgmed(\mathcal{C}) =\underset{y \in \mathbb{R}^2}{\operatorname{arg\,min}} \sum_{i=1}^{|\mathcal{C}|} w_i^p \left \| x_i-y \right \|_2.
\end{equation}
It can be  calculated using the Weiszfeld's algorithm \cite{Weiszfeld2008}, which is a form of iteratively re-weighted least squares \cite{daubechies2010iteratively} which iteratively computes
\begin{equation}
\left. y_{t+1}=\left( \sum_{j=1}^{|\mathcal{C}|} \frac{w_j^p x_j}{\| x_j - y_t \|} \right) \right/ \left( \sum_{j=1}^{|\mathcal{C}|} \frac{w_j^p}{\| x_j - y_t \|} \right),
\end{equation}
where again $p$ is a parameter of the method.



\subsection{Local Optimization}
One of the major improvements of the original RANSAC method, in terms of accuracy, is to include an intermediate step every time a new best hypothesis has been found. This step, known as the local optimization step, achieves improved results by using the fact that models with good enough inlier supports are probably not distant from the true underlying model. Therefore, by taking non-minimal samples from the inliers of this new best model followed by a greedy strategy to refine the results, the LO-RANSAC algorithm \cite{chum_locally_2003} and its variants \cite{lebeda_fixing_2012, raguram_usac_2013} improve systematically over RANSAC. 

The local optimization step is shown in Algorithm~\ref{alg:LO}. 
The method begins by taking non-minimal samples of the inliers of the new best hypothesis found, and computes the model by performing a least squares fit (lines 10-11)
. Then, it computes the inliers of this model for the whole dataset using a larger distance threshold (line 12) 
and uses these inliers to calculate a second model by least squares fitting (line 15). This second model is later refined in an iterative fashion, by progressively shrinking the distance threshold until its original value is attained (lines 16-21). The refined model with the largest amount of inliers is then returned.

In practice, if an all-inlier sample is picked during a RANSAC iteration, this method obtains excellent approximations to the optimal solution. However, since RANSAC only keeps the best model defined by its inlier support, the last may not necessarily be the optimal model, as a better model may have appeared during these inner iterations and was subsequently discarded. This suggests that \RANSAAC may significantly benefit from this procedure, and it turns out that it does. Therefore, saving and returning these intermediate models is the only extra work needed to adapt this method to \RANSAAC, as seen by the lines with pink background in Algorithm~\ref{alg:LO}.

\begin{algorithm}[htpb]  
 \caption{Local Optimization}
 \label{alg:LO}
 \begin{algorithmic}[1]
 \algrenewcommand\algorithmicindent{0.8em}%
\Require $X_1, X_2 \in \mathbb{N}^2 \times N$: $N$ matching points, \ \ \ $Xi_1, Xi_2 \in \mathbb{N}^2 \times K$: $K$ inlier matches, $\delta_d \in \mathbb{R}$: RANSAC distance threshold, $s_{is} \in \mathbb{N}$: size of inner sample, $m_{\delta_d} \in \mathbb{R}$: threshold multiplier, $reps \in \mathbb{N}$: inner sample repetitions, $lsIters \in \mathbb{N}$: least squares iterations, 
\If{$K < 2 s_{is}$}
	\State{\textbf{return} 0}
\EndIf
\State{$\Delta_{\delta_d} \leftarrow (m_{\delta_d} \cdot \delta_d - \delta_d)/lsIters$}
\State{$maxScore \leftarrow K$}
\State{$bestModel \leftarrow []$}
\CSTATE{$models \leftarrow []$}
\CSTATE{$weights \leftarrow []$}
\For{$it=1$ to $reps$}
	\State{$Xss_{1}, Xss_{2} \leftarrow GetSample(Xi_1, Xi_2, s_{is})$} 
    \State{$h_{nonMSS} \leftarrow GetModelLS(Xss_{1}, Xss_{2})$} 
    \State{$Xii_1,Xii_2 \leftarrow EvalH(h_{nonMSS}, X_1, X_2, \delta_d \cdot m_{\delta_d})$}
    \CSTATE{$Add(models, h_{nonMSS})$} \Comment{Save computed model\,\,}
    \CSTATE{$Add(weights, \#Xii_1)$} \Comment{and its weight\,\,}
    \State{$h_{RLS} \leftarrow GetModelLS(Xii_{1}, Xii_{2})$}
    \For{$j=1$ to $lsIters$}
	    \State{$Xii_1,\!Xii_2\!\leftarrow\!EvalH(h_{RLS}, X_1, X_2, \delta_d \cdot m_{\delta_d}\!-\!j \cdot \Delta_{\delta_d})$}
	    \CSTATE{$Add(models, h_{RLS})$} \Comment{Save computed model\,\,}
    	\CSTATE{$Add(weights, \#Xii_1)$} \Comment{and its weight\,\,}
	    \State{$h_{RLS} \leftarrow GetModelLS(Xii_{1}, Xii_{2})$} 
    \EndFor
    \State{$Xii_1, Xii_2 \leftarrow EvalH(h_{RLS}, X_1, X_2, \delta_d)$}
    \CSTATE{$Add(models, h_{RLS})$} \Comment{Save computed model\,\,}
   	\CSTATE{$Add(weights, \#Xii_1)$} \Comment{and its weight\,\,}
    \If{$\#Xii_1 > maxScore$}
    	\State{$maxScore \leftarrow \#Xii_1$}
        \State{$bestModel \leftarrow h_{RLS}$}
    \EndIf
\EndFor
\State{$\textbf{return } maxScore, bestModel, ${\colorbox{red!15!white}{$models, weights$}}}
 \end{algorithmic}%
\end{algorithm}%

As in its original formulation, the local optimization is executed every time a new best hypothesis is found during RANSAC iterations, working as an ``algorithm within an algorithm''. This means that the local optimization step does not update RANSAC's internal \emph{best-so-far} variables ($maxScore$ and $ransacRec$), but has its own. In the locally optimized version of \RANSAAC (i.e. LO-\RANSAAC), instead of aggregating all generated hypotheses, we only aggregate hypotheses obtained from the local optimization step, since these models are likely to be closer to the ground truth. We evaluated the inclusion of all other generated models in the aggregation step and observed that the results were slightly less accurate. Furthermore, excluding these hypotheses in the aggregation reduces the processing time. 



The final \RANSAAC algorithm and its locally optimized version add a few lines to the traditional RANSAC method and an almost negligible computational cost. Algorithm \ref{alg:aransacWithLO} shows RANSAC together with both \RANSAAC and LO-\RANSAAC approaches, coded in colors. Each variant is understood by its identifying color. The original RANSAC is coded with white background. The \RANSAAC method is obtained by including the pink and grey lines. The LO-\RANSAAC procedure includes green and grey lines.  As seen within the lines with pink background,  \RANSAAC first verifies for each iteration if the hypothesis is acceptable (i.e. that the amount of inliers is larger than the MSS), and then applies it to the source points while storing its scores as weights. The LO procedure, coded in green background, just performs the local optimization step when a new best model is found, and stores its results. Finally, after iterating, if no acceptable transform was found, it returns the traditional RANSAC result. However, if several hypotheses were considered valid, the computed projected points are aggregated using their weights and the algorithm returns the model that best fits these estimates.

%

\begin{algorithm}  
 \caption{Computing a 2D transformation using the (LO) \RANSAAC algorithm.
 }
 \label{alg:aransacWithLO}
 \begin{algorithmic}[1]
\Require $X_1, X_2 \in \mathbb{N}^2 \times N$: $N$ matching points,  $minSamples \in \mathbb{N}$: minimum amount of matches to compute the transform,
$srcPts \in \mathbb{N}^2\times K$: vector with $K=minSamples$ input points,
$\delta_d \in \mathbb{R}$: RANSAC distance threshold, 
$iters \in \mathbb{N}$: amount of iterations, $p \in \mathbb{R}$: aggregation parameter.
\State $maxScore \leftarrow 0$
\For{$it=1$ to $iters$}
	\State{$Xss_1, Xss_2 \leftarrow GetSample(X_1, X_2, minSamples)$} 
    \State{$h \leftarrow GenerateHypothesis(Xss_1, Xss_2)$}
    \State{$Xi_1,Xi_2 \leftarrow EvalH(h, X_1, X_2, \delta	_d)$}
    \If{$\#Xi_1 > maxScore$}
    	\State{$maxScore \leftarrow \#Xi_1$}
        \State{$ransacRes \leftarrow h$}
        \CSTATETWO{\small{$models, weights \leftarrow LO(X_1, X_2, Xi_1,Xi_2, \delta_d)$}}
    	\CSTATETWO{$Add(dstPts_{LO}, Project(srcPts, models))$} 
        \CSTATETWO{$Add(weights_{LO}, weights$)} 
    \EndIf
    \CIf{$\#Xi_1 > minSamples$}
    	\CSTATE{$Add(dstPts, Project(srcPts, h))$}
    	\CSTATE{$Add(weights, \#Xi_1)$}
    \CEndIf
\EndFor
\CSTATETWO{$dstPts \leftarrow dstPts_{LO}$} 
\CSTATETWO{$Ws \leftarrow weights_{LO}$}
\CIfThree{$\#dstPts > 0$}
	\CSTATETHREE $resPts \leftarrow Aggregate(dstPts, Ws, p)$ \Comment{{\small See \ref{subsec:aggregation}\,\,}}
    \CSTATETHREE{\textbf{return} $GenerateHypothesis(srcPts, resPts)$}
\CEndIfThree
	\State{\textbf{return} $ransacRes$}
 \end{algorithmic}%
\end{algorithm}%
\vspace{-3mm}

\section{Experiments}
\label{sec:evaluation}
We compared the proposed method and its variants (LO, $wmean$, $wgmed$) with several state-of-the-art approaches both qualitative and quantitatively. Two transformation types were considered, namely the projective transform estimated using the DLT algorithm \cite{Sutherland_74}, and a homographic model that allows distortions on both images. The latter was estimated using the $H_5$ method of Kukelova et al. \cite{Kukelova_2015}, which, given five matches, computes a homography and one radial distortion coefficient per image (using the division model and assuming the distortion center is on the center of the image). Since homographies are, along with fundamental matrices, the most frequent global model estimated in computer vision, we focus our evaluation on it. Nevertheless, we show results proving that the proposed approach also works on the other transform types as well, and could be potentially generalized to any parametric model obtained from points provided the possibility of averaging those models is available. 

Matlab implementations of \RANSAAC and all its variants are available at \url{http://dev.ipol.im/~mrais/ransaac/}.

\subsection{Qualitative Evaluation: Paintings Registration with Projective Transformations}
\label{sec:qualitativeProjecive}
\begin{figure*}
\captionsetup[subfigure]{labelformat=empty}
\centering
\begin{subfigure}{0.33\textwidth}
  \centering
  \includegraphics[width=1\linewidth]{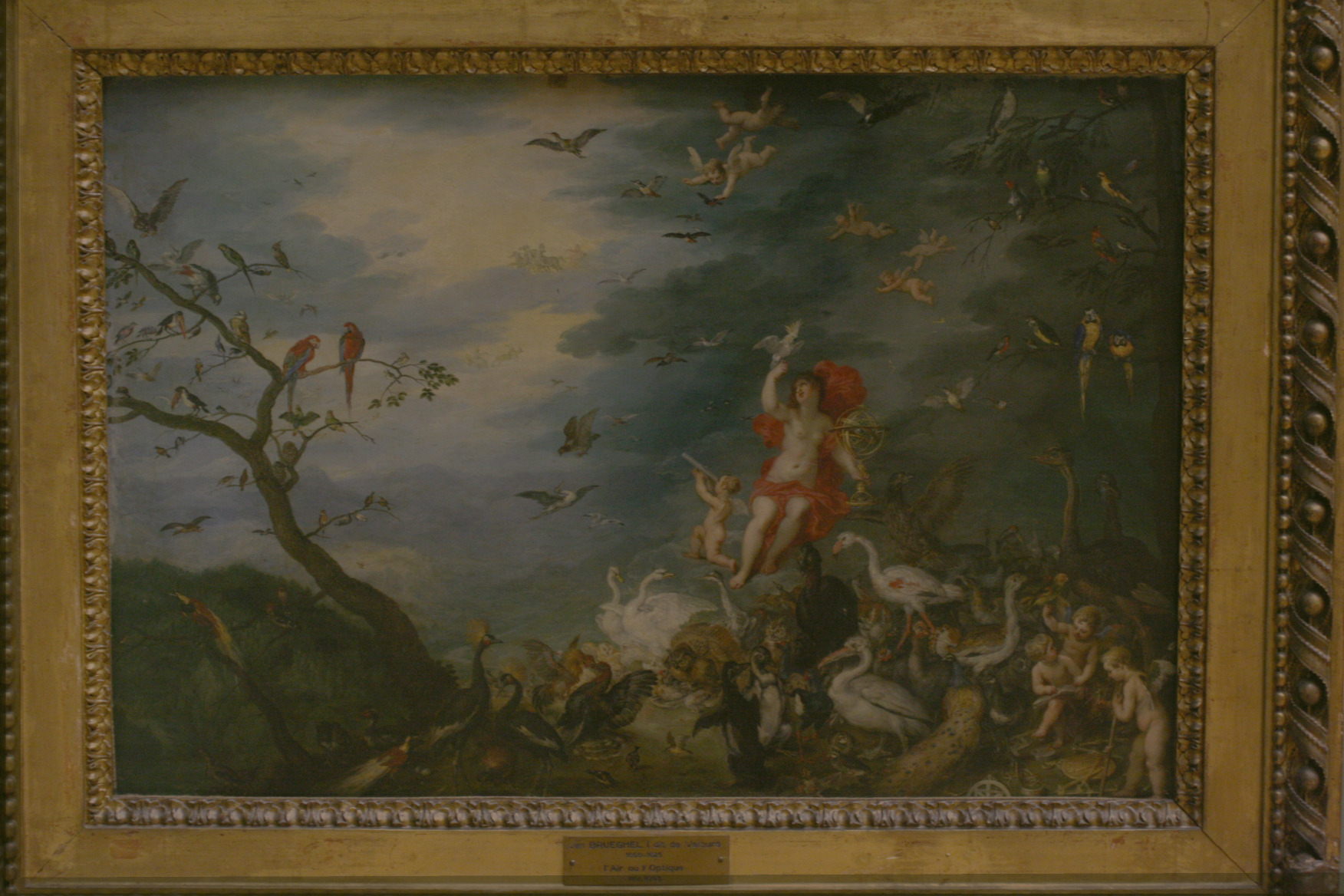}
\end{subfigure}\hspace{.1em}%
\begin{subfigure}{0.33\textwidth}
  \centering
  \includegraphics[width=1\linewidth]{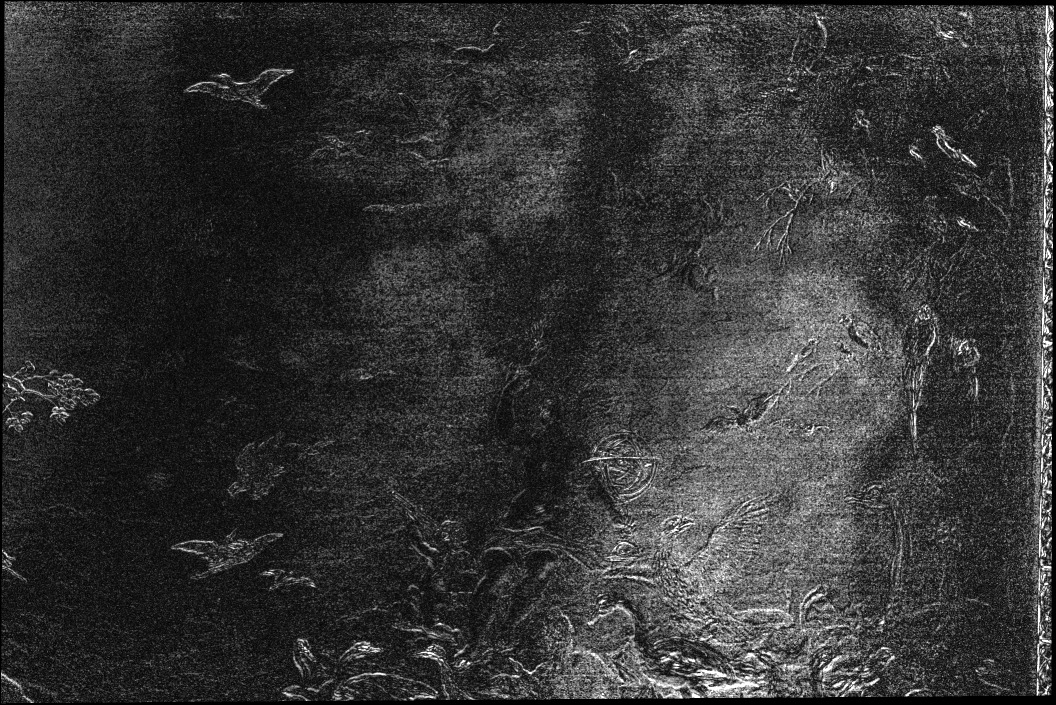}
\end{subfigure}\hspace{.1em}%
\begin{subfigure}{0.33\textwidth}
  \centering
  \includegraphics[width=\linewidth]{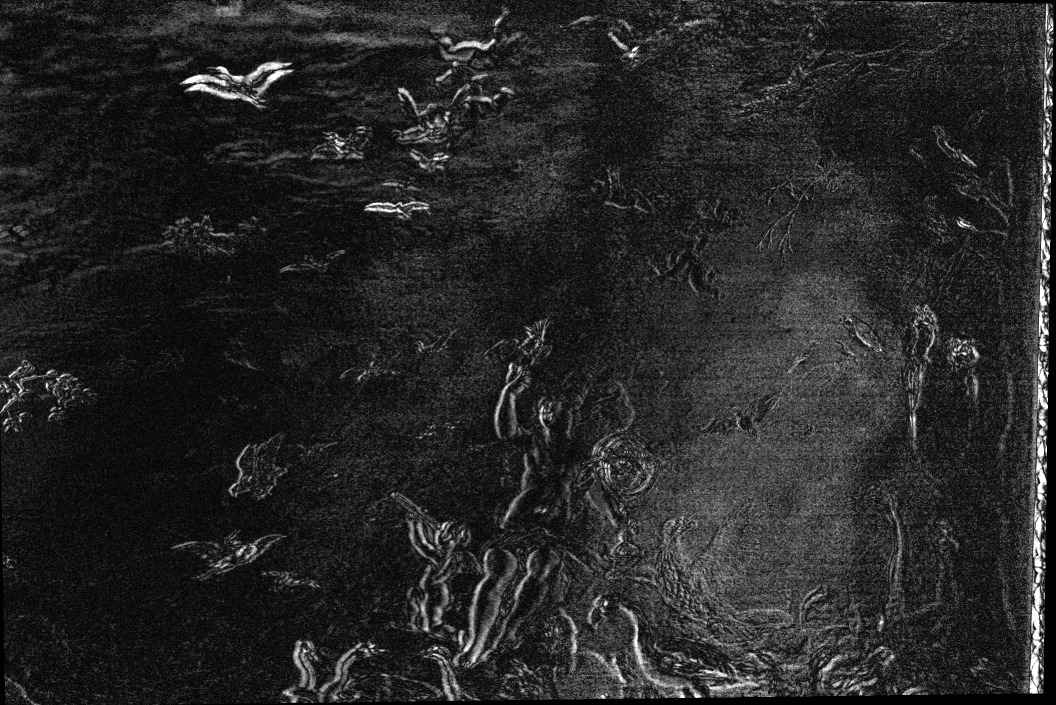}
\end{subfigure}

\vspace{.1em}%
\begin{subfigure}{0.33\textwidth}
  \centering
  \includegraphics[width=\textwidth]{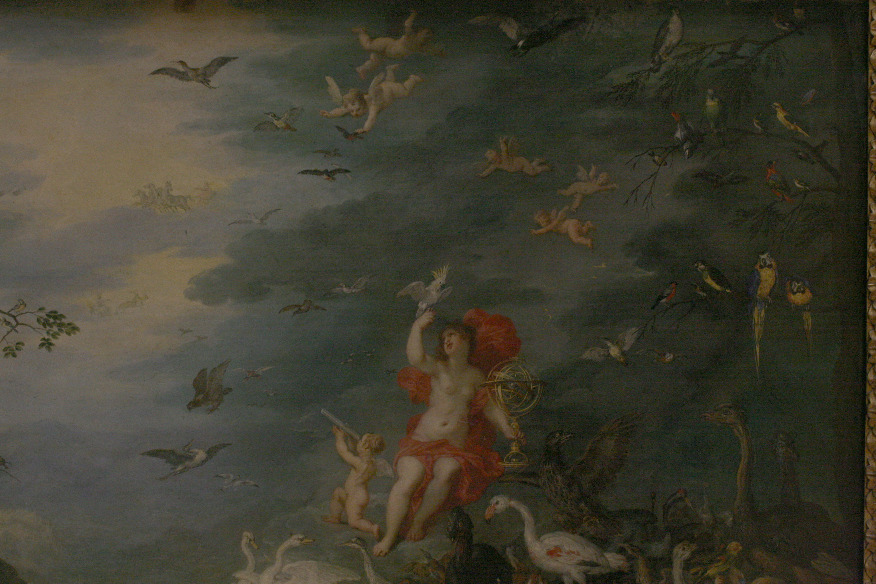}
\end{subfigure}\hspace{.1em}%
\begin{subfigure}{0.33\textwidth}
  \centering
  \includegraphics[width=\linewidth]{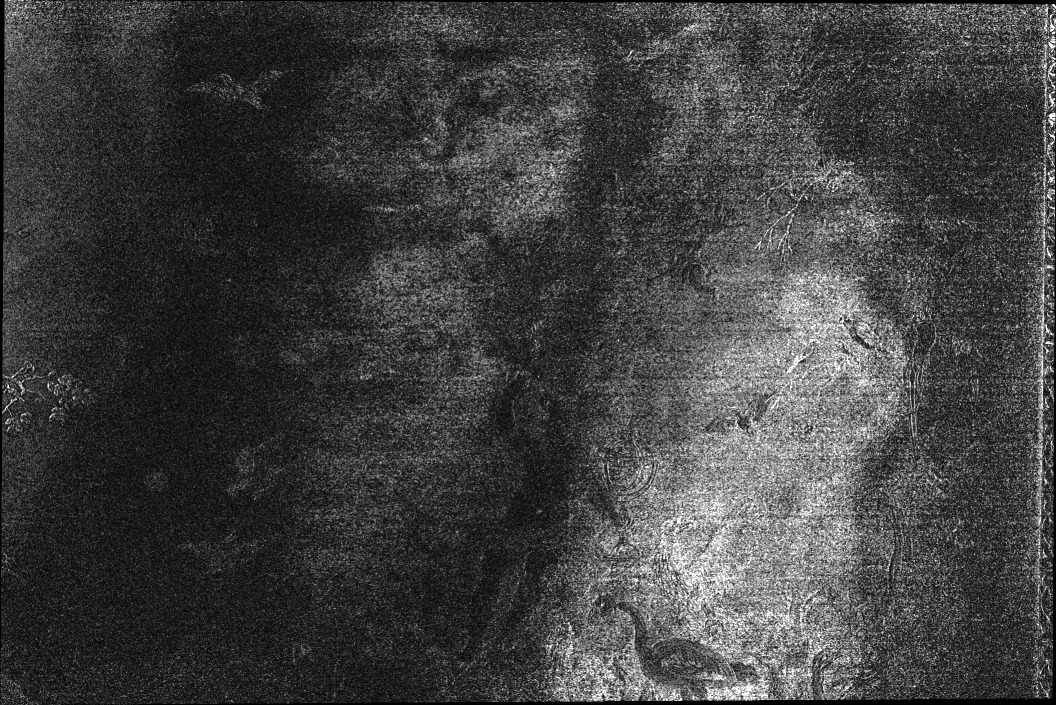}
\end{subfigure}\hspace{.1em}%
\begin{subfigure}{0.33\textwidth}
  \centering
  \includegraphics[width=\textwidth]{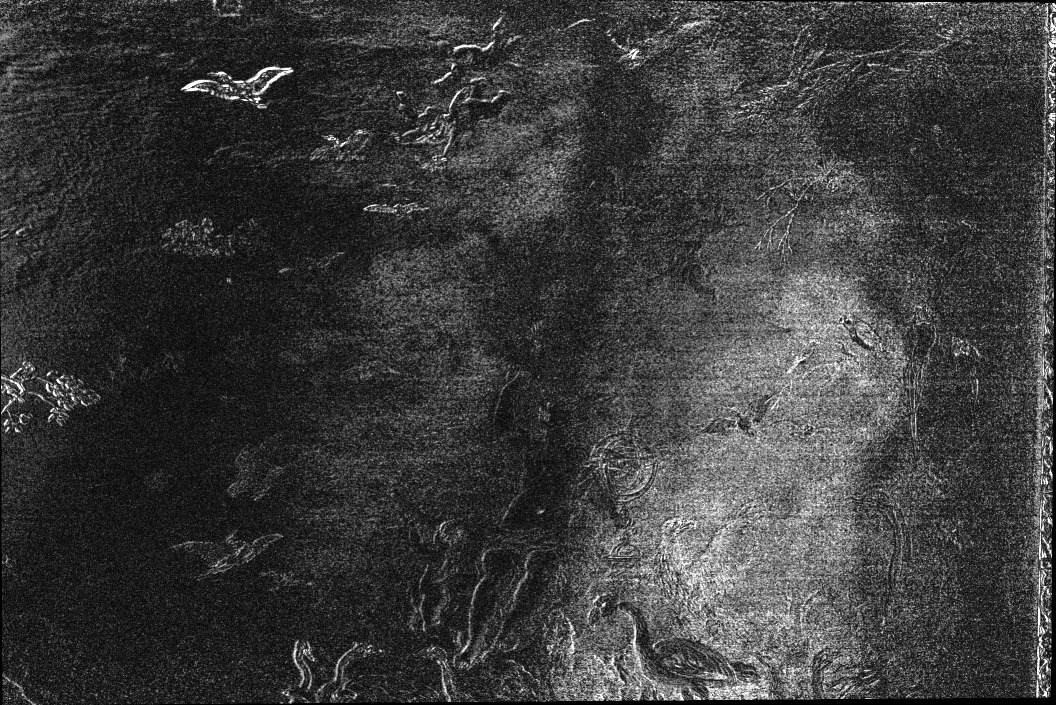}
\end{subfigure}
  \caption{\small{RANSAC (top) and \RANSAAC (bottom) registration results. Dynamic ranges were stretched to highlight differences. Left: input images. Center: registration difference using input images. Right: registration difference by adding measurement noise of $\sigma=5$ to the keypoint positions on both images in pixels. Edges due to misalignments can be perceived in the difference image, particularly by using the traditional RANSAC method. Painting: L'Air, ou L'Optique. Jan I BRUEGHEL, 1621. Louvre Museum.}}
\label{fig:Painting1}
\end{figure*}

We first evaluated both RANSAC and \RANSAAC with weighted mean aggregation on pictures of paintings taken in a museum under poor lighting conditions, which resulted in considerable  noise visible on the output images. The objective was then to register each photograph of a painting detail taken from a shorter range, with the whole painting \cite{Buades_2015}. By using SIFT features and matching its descriptors, we proceeded to register both images using both methods. An example can be seen in Fig.~\ref{fig:Painting1}. The camera distortion was corrected beforehand so  the images are related by a pure homography. After registering both images, the resulting difference is mainly noise, however \RANSAAC  performs better, as evidenced in the flamingo on the bottom part of the painting (middle column of Fig.~\ref{fig:Painting1}). When measurement noise is added to the input points on both images ($\sigma\!=\!5$), the difference in the registration performance becomes much more evident.  Edges due to misalignments appear all over the difference image in RANSAC, but are dimmer on the \RANSAAC results (right row of Fig.~\ref{fig:Painting1}).

\begin{figure}
\centering
\includegraphics[width=.9\linewidth]{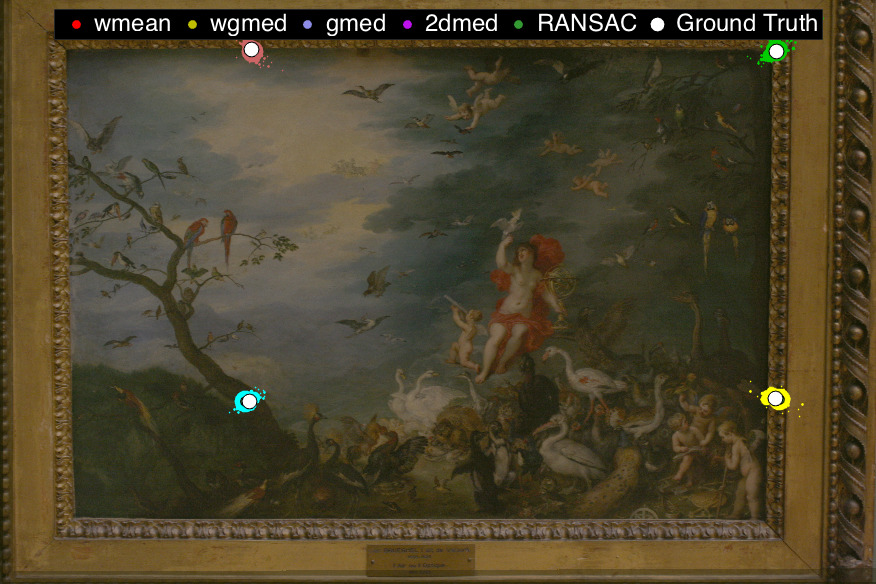}
\caption{\small{Resulting point distributions for 1000 iterations.}}
\label{fig:PointDistributions}
\vspace{-3mm}
\end{figure}

\begin{figure*}
\centering
\includegraphics[trim={30 100 30 0}, clip, width=.48\linewidth, height=4.5cm]{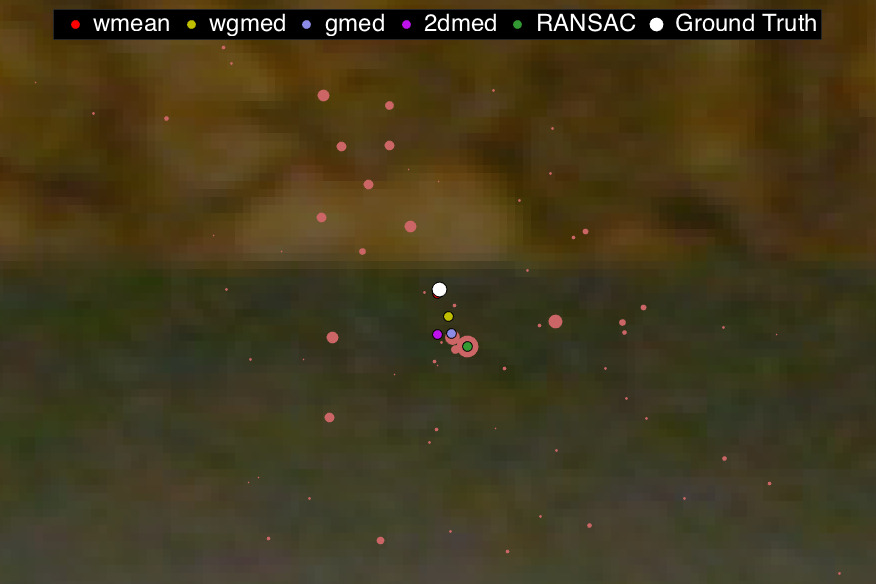}\hspace{.1em}%
\includegraphics[trim={30 100 30 0}, clip,width=.48\textwidth, height=4.5cm]{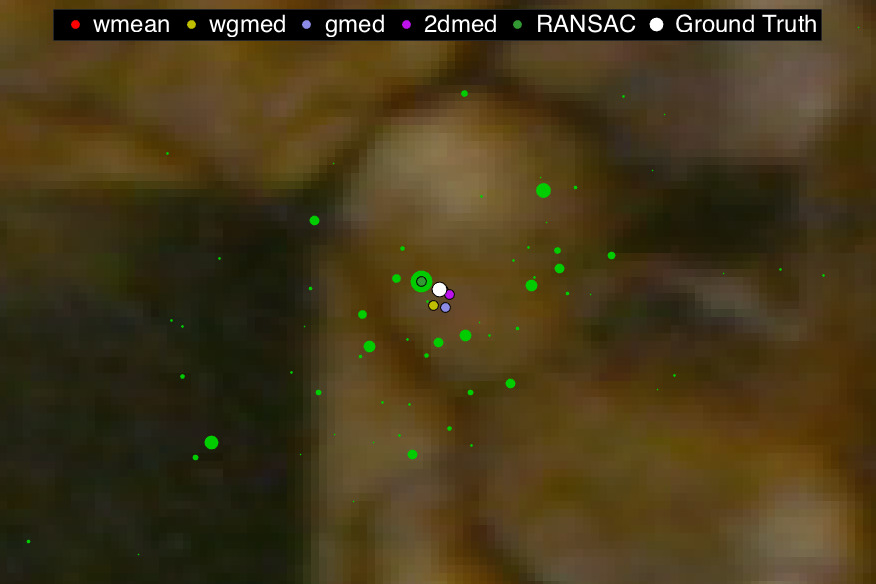}

\vspace{.1em}%
\includegraphics[trim={30 100 30 0}, clip,width=.48\linewidth, height=4.5cm]{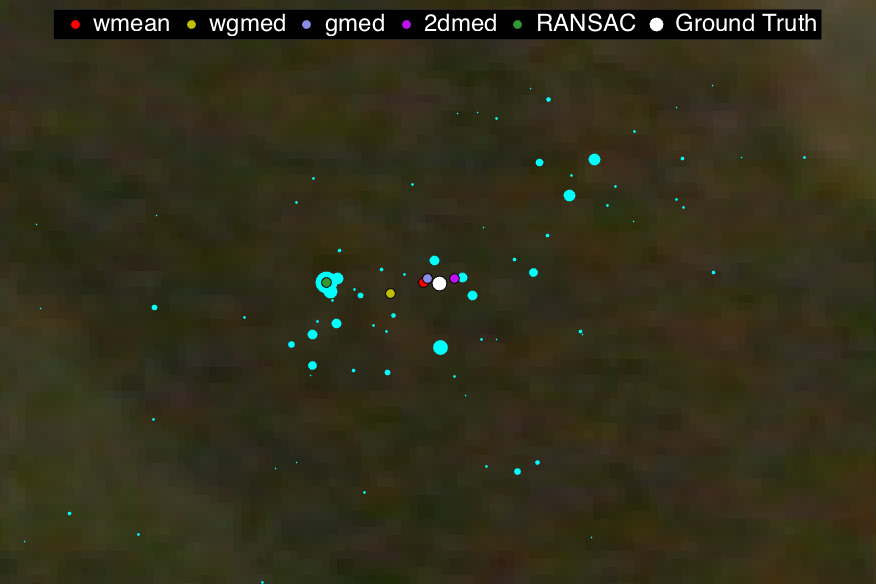}\hspace{.1em}%
\includegraphics[trim={30 100 30 0}, clip,width=.48\textwidth, height=4.5cm]{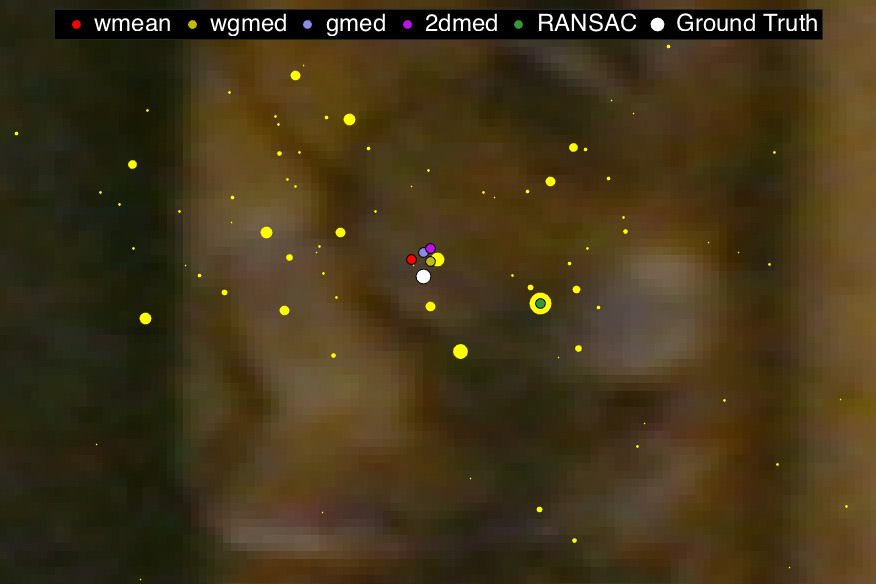}
\caption{Point distribution of each corner of Fig.~\ref{fig:PointDistributions} resized according to their weights using 1000 iterations. Points with low weights were excluded for visualization purposes. $wmean$ and $wgmed$ are the weighted mean and the weighted geometric median aggregation. $gmed$ and $2dmed$ represent aggregated results using (unweighted) geometric median and 1D median on both dimensions. The reader is advised to zoom in to see better the positions of the various estimates.}
\label{fig:exampleResultingPoints}
\end{figure*}

To get a better understanding of the behaviour of the method, figures \ref{fig:PointDistributions} and its zoomed in version \ref{fig:exampleResultingPoints} shows the point distribution with their sizes according to their obtained weights using $p=5$. As can be seen, the RANSAC algorithm always returns the model given by the points with the highest weight, ignoring every other estimation. By incorporating other weighted estimations and aggregating them, a closer point to the ground truth is obtained for every corner, therefore yielding a more precise homography.


\subsection{Quantitative Evaluation by a Simulated Ground Truth. Application 1: Estimating Projective Transforms.}
\label{sec:quantitativeEvaluation}
We built a valid ground truth, composed of point matches and a transform. In order to obtain a valid real-life transform, we used the SIFT algorithm, followed by the application of RANSAC with 100000 iterations between two images. To generate the matches, we randomly selected a fixed set of points from the first image as inliers and we matched them to the points obtained by applying  the ground truth transform to them. Finally, to incorporate outliers in the dataset, we randomly picked other points from the first image and matched them with random positions on the second image. Measurement noise up to $\sigma=5$ was added afterwards, since such high noise levels may appear in real-life applications. A possible reason for this is noisy point matches coming from higher scales of the Laplacian pyramid, which translate to large noisy values on the original scale. Another reason is due to an imperfect model selection that, by setting large values for $\delta_d$, accepts matching points far away from the selected model. For example, when searching for a planar homography on radially distorted images from the same planar scene.

On this dataset, we compared  several state-of-the-art methods and all variants of \RANSAAC. To measure the error, we averaged the symmetric transfer error in pixels for every inlier. If $\phi_{\theta}$ is the obtained model, $X_i$ is an inlier point in the first image, $Y_i$ its match and $K$ the amount of inliers, then the mean error $E$ is defined as 
\begin{equation}
E = \frac{1}{K}\sum_{i=1}^K \frac{\norm{\phi_{\theta}(X_i) - Y_i}_2 + \norm{\phi^{-1}_{\theta}(Y_i) - X_i}_2}{2},
\end{equation}
where $X_i$ and $Y_i$ represent the original input points, before adding the noise.

By varying the number of inliers, the percentage of outliers, the input noise and the number of iterations performed, we computed for each experiment the average error per experiment $\bar{E}$ and the standard deviations $\sigma_E$ over 100 evaluations. 

Note that for all presented results, the very same random samples for each iteration were used for RANSAC, LO-RANSAC and \RANSAAC, thus ensuring equal chance for those methods. We used the implementation of USAC available from the author \cite{raguram_usac_2013}.

We evaluated the performance of the proposed approach under distinct conditions. Since the noise level was known beforehand, the RANSAC $\delta_d$ parameter was set according to section \ref{sec:distanceParam}, and for \RANSAAC this parameter was fixed to $2*\delta_d$ due to the higher tolerance of the method with respect to this parameter. 
Unless explicitly mentioned, the parameter $p$ was set depending on the amount of iterations performed and on the used aggregation scheme, indicated by the results obtained in the following section.

\subsubsection{Aggregation Weights}
\label{subsec:aggWeightsComparison}
Because of outliers in the input data, the weighting of the hypotheses on the aggregation is mandatory. 
Increasing the value of the weight parameter $p$ leads to eliminate the influence of poor hypotheses, but a too high value will imply using too few transforms for the final aggregation. 

\begin{figure}
\centering
\captionsetup[subfigure]{labelformat=empty}
\begin{subfigure}{0.50\textwidth}
  \centering
  \includegraphics[width=.9\linewidth, height=4.2cm]{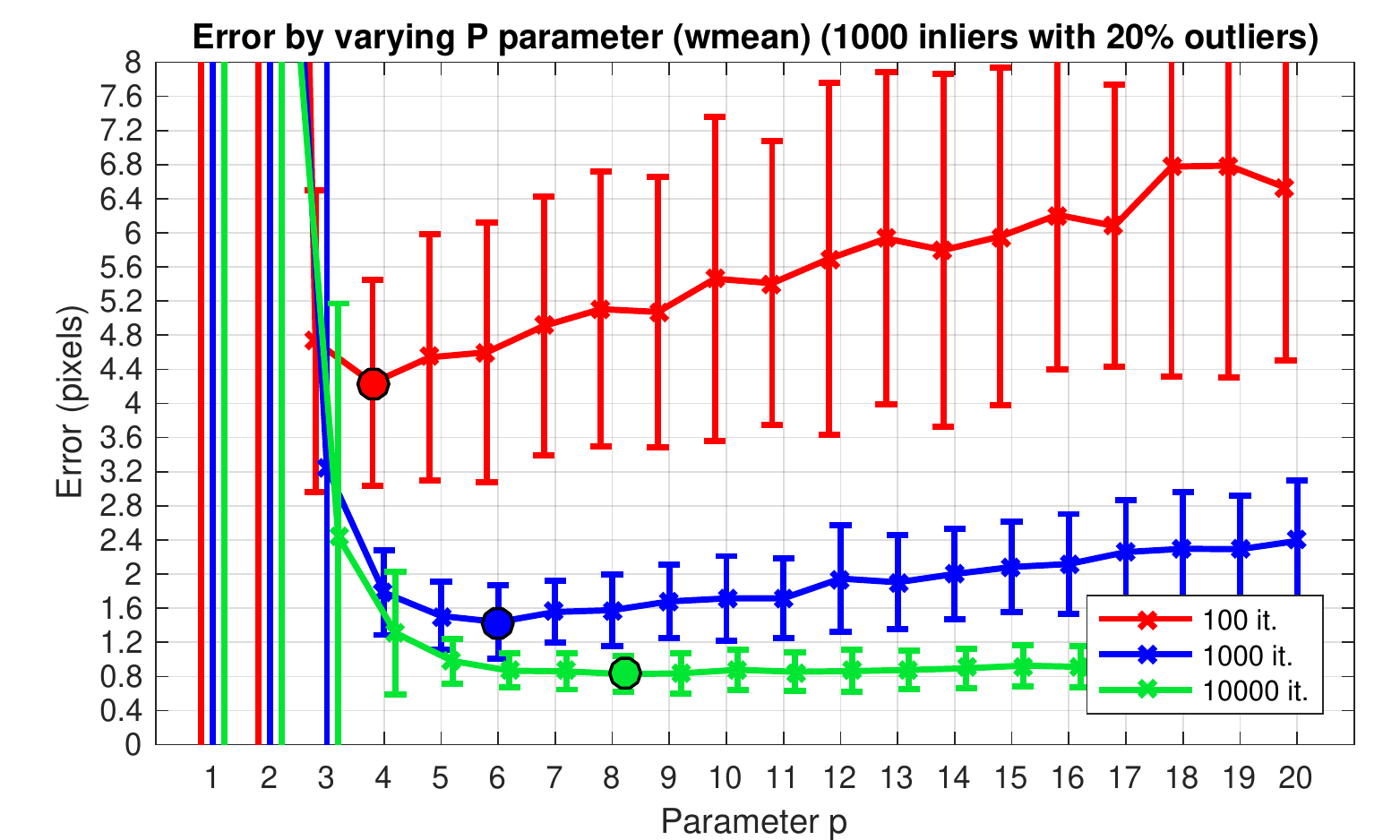}
\end{subfigure}
\begin{subfigure}{0.50\textwidth}
  \centering
  \includegraphics[width=.9\textwidth, height=4.2cm]{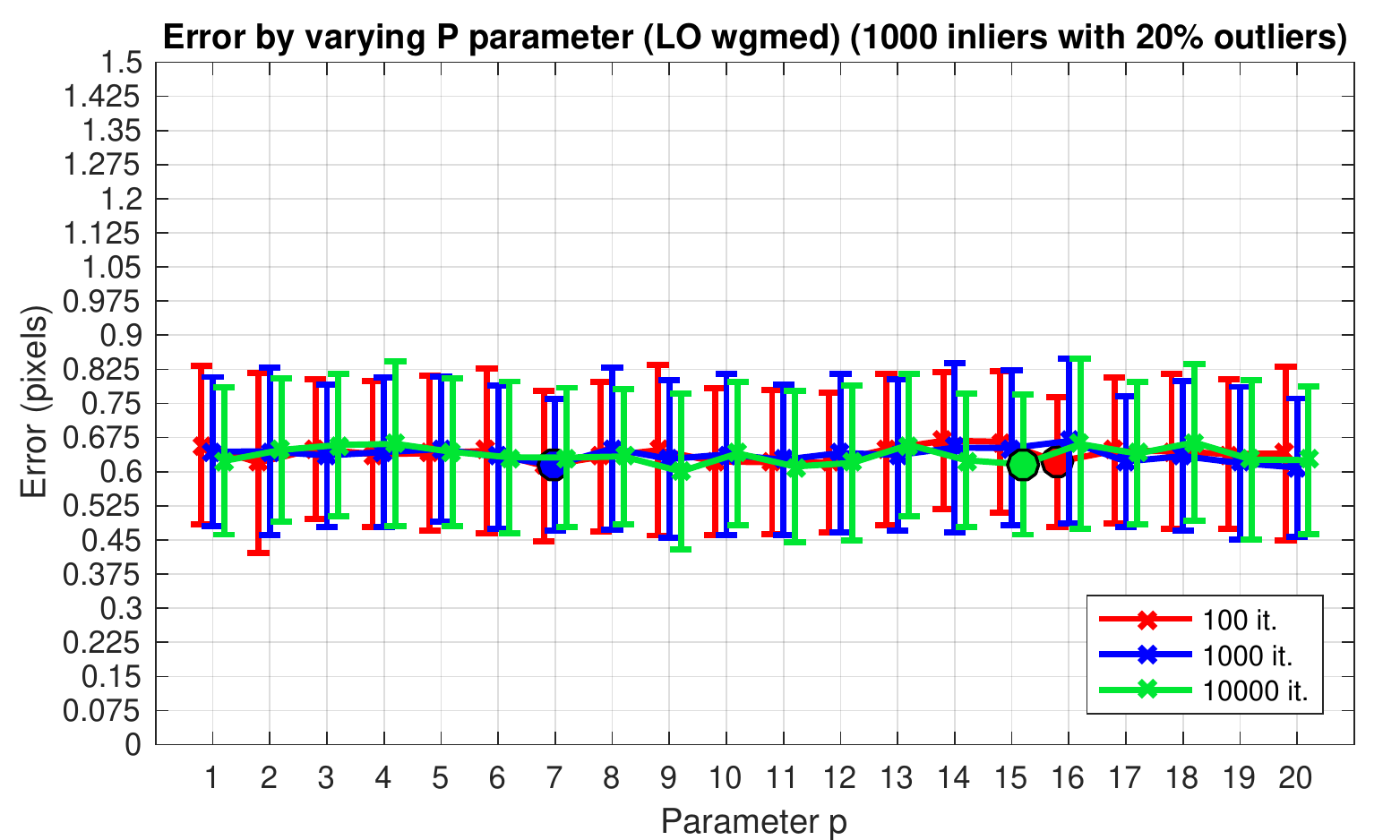}
\end{subfigure}


\caption{Error (in pixels) and std. dev. by varying the $p$ exponent, using 100, 1000 and 10000 iterations averaged over 100 realizations. Dots indicate lowest errors. \textbf{Top}: \RANSAAC with $wmean$. \textbf{Bottom}: LO-\RANSAAC with $wgmed$. Inlier/Outlier ratio: 1000/1250. Noise: $\sigma\!=\!5$.}
\label{fig:bestPExperiment}
\end{figure}

An experiment was performed to measure the impact of the parameter $p$ for \RANSAAC and its LO-\RANSAAC variant, shown in Fig.~\ref{fig:bestPExperiment}. For the case of \RANSAAC, it turns out that the value of $p$ depends on both the outlier percentage, but most importantly, on the number of drawn hypotheses. Indeed, as more hypothesis are drawn, they should be better discriminated, suggesting higher values of $p$. As soon as the value of $p$ is high enough so that it allows to correctly discriminate between inliers and outliers, then the resulting accuracy does not vary significantly. A similar trend was observed using $wgmed$ aggregation and therefore its results were not included. However, for the LO-\RANSAAC variant using $wgmed$ aggregation, varying the $p$ parameter does not affect the resulting accuracy. This can be explained in two ways. First, the models returned by the LO step have usually fewer outliers. Secondly, the geometric median is more robust to outliers than the mean, therefore not requiring to completely avoid them in the aggregation. When using $wmean$ aggregation, results were similar as in the standard \RANSAAC with low values of $p$. However, the curves remained almost constant later, as with the $wgmed$.

\subsubsection{Accuracy evolution along iterations}
\label{subsec:evolutionAlongIterations}
\begin{figure}
\centering
\hspace{-2.5mm}
\captionsetup[subfigure]{labelformat=empty}
\begin{subfigure}{.45\textwidth}
  \centering
  \includegraphics[width=\linewidth, height=4.2cm]{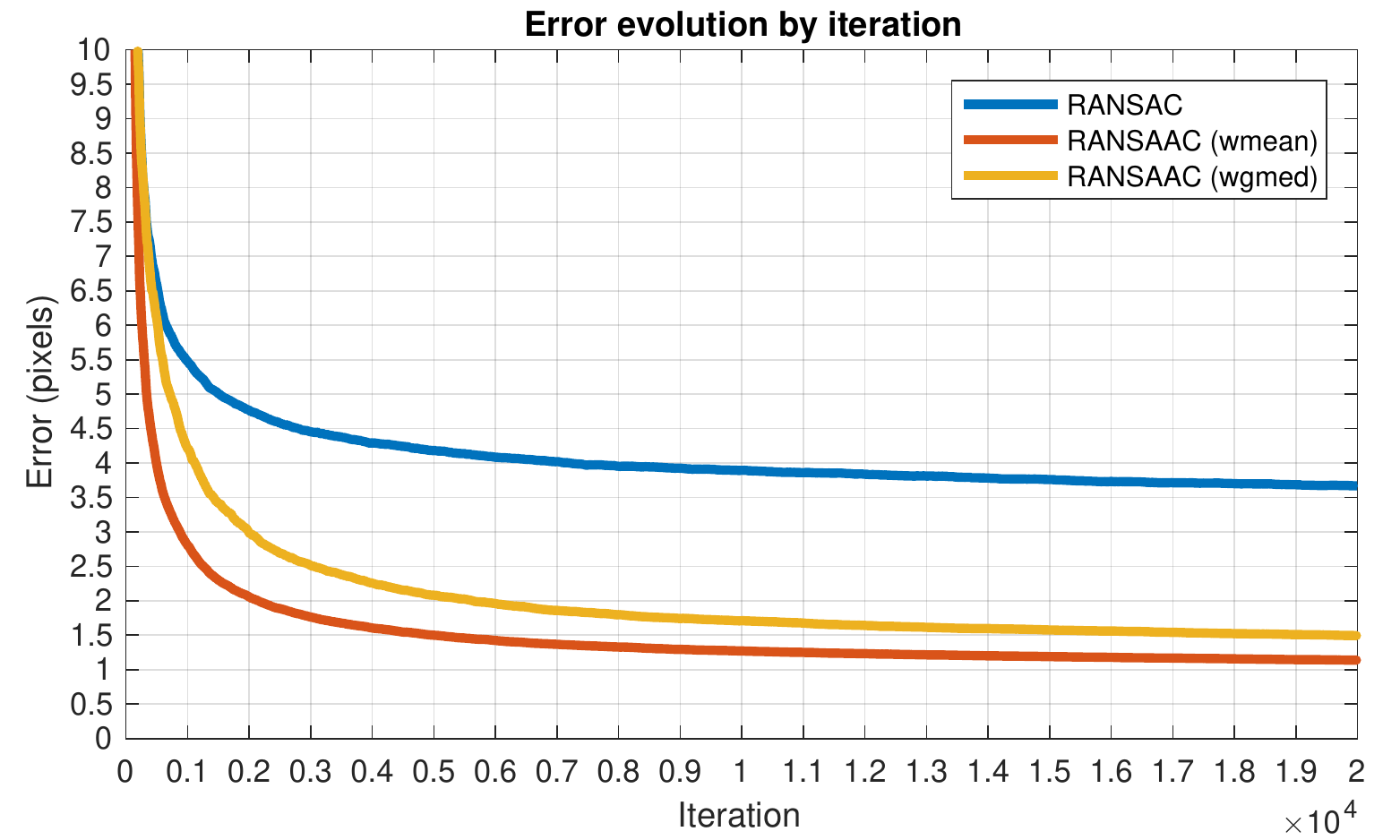}
\end{subfigure}
\begin{subfigure}{.45\textwidth}
  \centering
  \includegraphics[width=\textwidth, height=4.2cm]{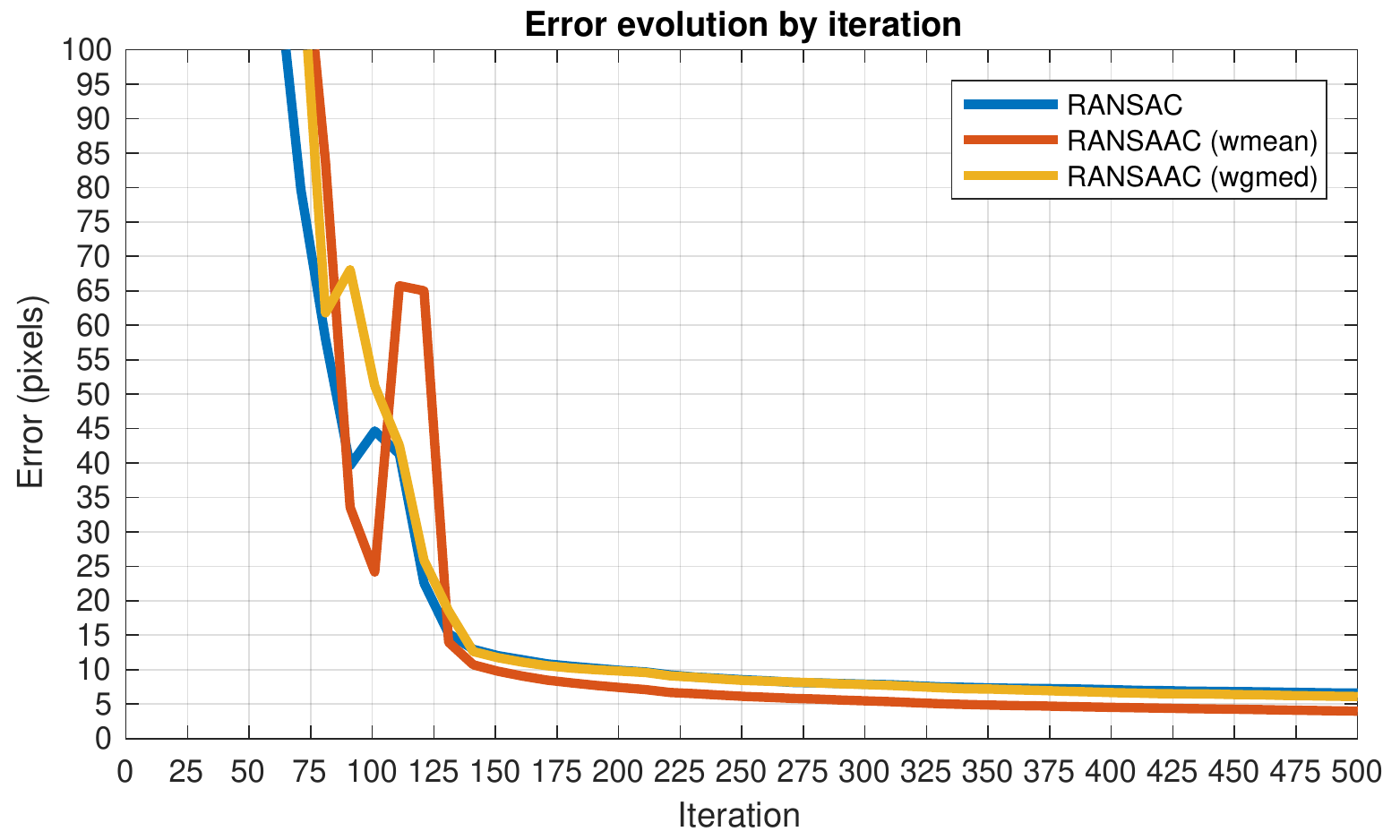}
\end{subfigure}
\caption{Comparison of the error by iteration (in pixels) between RANSAC and the proposed algorithm using both aggregation methods averaged over 1000 realizations. Noise: $\sigma=5$. 100 inliers and 100 outliers. \textbf{Top}: Up to 20000 iterations. \textbf{Bottom}: First 500 iterations.}
\label{fig:partialExperiment}
\end{figure}

To give insight about the evolution of the error under RANSAC and under the proposed approach, their accuracy was computed while iterating. To this end, we simulated input points from a predefined real homography (i.e. computed from two images of a landscape), corrupted them with white Gaussian Noise of $\sigma=5$ and added 50\% outliers. Using this input, we ran 20000 iterations of RANSAC and our method and calculated the error for every iteration (for \RANSAAC, this implied performing aggregation). Finally, we computed the average error per iteration from 1000 repetitions of the experiment. In Fig.~\ref{fig:partialExperiment} left, we observe how RANSAC does not only converge slower comparing it with both aggregation schemes of \RANSAAC, but also that it never reaches their accuracy. 
However, when observing the first 500 iterations, RANSAC usually achieves better results at the beginning, while \RANSAAC usually requires more \emph{all-inlier} samples to produce more accurate results. This is expected, since \RANSAAC gains from aggregating all generated samples, thus only having a single ``good'' hypothesis and several ``bad'' ones will always benefit a method such as RANSAC that just considers the single hypothesis with the highest obtained score. However, as more \emph{all-inlier} hypotheses are sampled, \RANSAAC starts improving over traditional RANSAC (in the figure this happens at 125 iterations).

\subsubsection{\RANSAAC variants evaluation}
\label{sec:loEvaluation}
We evaluated both aggregation variants for \RANSAAC with and without local optimization, by varying the number of inliers, the percentage of outliers, noise level, and the number of  iterations performed. Table \ref{tab:RANSAACvsLORANSAAC} shows results under three different noise levels ($\sigma=0.5, 2$ or $5$). For each method and noise level, both 1000 and 10000 iterations were tested, each row corresponding to situations presenting either 0\%, 5\%, 20\% or 50\% outliers respectively. 

Several conclusions could be drawn from these results. As can be seen, by using the local optimization procedure, every aggregated method improved, and in every case, the best performing aggregation scheme  was the weighted geometric median. Interestingly, this did not occur when no local optimization steps were performed. The reason behind this is evident, as both the outlier ratio and the number of inliers increase, the probability of obtaining good models by random sampling becomes lower, therefore averaging these not-so-close-to-the-optimal models will achieve better results than taking their median, which will be far away from the optimal solution. However, when better (i.e. closer to the optimal solution) hypotheses are available, then computing their median is not only more accurate but also drastically reduces the possible bias caused by including outliers in the estimation. Indeed, performing LO steps provides excellent models, which justifies why $wgmed$ aggregation improves over $wmean$. 

It is remarkable how the $wgmed$ aggregation benefits from local optimization, evidenced by the increase in accuracy up to an order of magnitude for 1000 iterations, 1000 inliers, 50\% outliers and $\sigma\!=\!5$. Finally, performing more RANSAC iterations when using local optimization does not improve the resulting accuracy. Indeed, results could become slightly worse when performing more than enough iterations since there are more possible bad hypotheses that have to be correctly discriminated. 
In conclusion, adding local optimization to \RANSAAC does not only improves results but also enables the method to perform fewer iterations, making it less computationally demanding. 

To sum up our findings, LO-\RANSAAC using \emph{wgmed} is our recommended algorithm whenever local optimization is applicable. When this is not possible, the best \RANSAAC results are obtained by using the \emph{wmean} aggregation. An example of this is shown in section \ref{sec:HomoPlusDistortion}.

\begin{table}
\caption{Average errors for the different versions of \RANSAAC using both proposed aggregation schemes. For noises $\sigma=0.5, 2$ and $5$, each method was evaluated with 100 and 1000 inliers using both 1000 and 10000 iterations. The four errors (rows) represent four outlier ratios: 0\%, 5\%, 20\% and 50\%. Bold denotes the best performers.}
\label{tab:RANSAACvsLORANSAAC}
\centering
{\footnotesize{
\def\arraystretch{.85}%
\begin{tabular}{c|c|c|c|c|c|c|c|c|}
$\sigma$ & \multicolumn{2}{ c| }{wmean} & \multicolumn{2}{ c| }{LO (wmean)} & \multicolumn{2}{ c| }{wgmed} & \multicolumn{2}{ c| }{LO (wgmed)}\\ \hline 
 & 1k & 10k  & 1k & 10k  & 1k & 10k  & 1k & 10k \\ \hline 
 & \multicolumn{8}{ |c| }{100 inliers} \\ \hline
\multirow{4}{*}{0.5} & $0.23$ & $0.21$ & $0.20$ & $\textbf{0.19}$ & $0.25$ & $0.23$ & $\textbf{0.19}$ & $\textbf{0.20}$\\ 
 & $0.24$ & $0.21$ & $0.20$ & $\textbf{0.20}$ & $0.27$ & $0.23$ & $\textbf{0.20}$ & $\textbf{0.20}$\\ 
 & $0.26$ & $0.21$ & $0.21$ & $\textbf{0.20}$ & $0.29$ & $0.23$ & $0.21$ & $\textbf{0.20}$\\ 
 & $0.45$ & $0.26$ & $0.21$ & $\textbf{0.21}$ & $0.61$ & $0.30$ & $\textbf{0.21}$ & $\textbf{0.21}$\\ 
\hline 
\multirow{4}{*}{2} & $0.89$ & $0.87$ & $0.75$ & $\textbf{0.81}$ & $1.02$ & $0.92$ & $\textbf{0.75}$ & $0.81$\\ 
 & $0.91$ & $0.86$ & $0.76$ & $\textbf{0.81}$ & $1.01$ & $0.91$ & $\textbf{0.76}$ & $0.81$\\ 
 & $1.03$ & $0.89$ & $0.83$ & $\textbf{0.83}$ & $1.21$ & $0.97$ & $0.83$ & $\textbf{0.82}$\\ 
 & $1.76$ & $1.03$ & $0.84$ & $0.82$ & $2.39$ & $1.23$ & $0.82$ & $\textbf{0.81}$\\ 
\hline 
\multirow{4}{*}{5} & $2.20$ & $2.07$ & $1.90$ & $\textbf{1.90}$ & $2.42$ & $2.29$ & $\textbf{1.88}$ & $1.91$\\ 
 & $2.39$ & $2.22$ & $2.02$ & $\textbf{2.07}$ & $2.72$ & $2.40$ & $\textbf{2.01}$ & $2.06$\\ 
 & $2.58$ & $2.25$ & $2.05$ & $\textbf{2.01}$ & $2.95$ & $2.47$ & $\textbf{2.03}$ & $\textbf{2.03}$\\ 
 & $4.60$ & $2.47$ & $2.01$ & $\textbf{1.94}$ & $6.00$ & $2.98$ & $1.98$ & $\textbf{1.97}$\\ 
\hline 
 & \multicolumn{8}{ |c| }{1000 inliers} \\ \hline
\multirow{4}{*}{0.5} & $0.11$ & $0.07$ & $0.07$ & $\textbf{0.07}$ & $0.15$ & $0.09$ & $0.07$ & $\textbf{0.06}$\\ 
 & $0.12$ & $0.07$ & $0.07$ & $\textbf{0.07}$ & $0.16$ & $0.09$ & $\textbf{0.06}$ & $\textbf{0.06}$\\ 
 & $0.18$ & $0.08$ & $0.07$ & $\textbf{0.07}$ & $0.22$ & $0.10$ & $\textbf{0.06}$ & $\textbf{0.06}$\\ 
 & $0.42$ & $0.15$ & $0.08$ & $\textbf{0.07}$ & $0.54$ & $0.22$ & $0.07$ & $\textbf{0.06}$\\ 
\hline 
\multirow{4}{*}{2} & $0.47$ & $0.30$ & $0.27$ & $0.27$ & $0.57$ & $0.35$ & $\textbf{0.24}$ & $0.26$\\ 
 & $0.52$ & $0.31$ & $0.28$ & $0.28$ & $0.64$ & $0.36$ & $\textbf{0.26}$ & $\textbf{0.26}$\\ 
 & $0.67$ & $0.33$ & $0.28$ & $0.27$ & $0.92$ & $0.40$ & $\textbf{0.25}$ & $\textbf{0.25}$\\ 
 & $1.68$ & $0.61$ & $0.32$ & $0.28$ & $2.21$ & $0.83$ & $0.26$ & $\textbf{0.25}$\\ 
\hline 
\multirow{4}{*}{5} & $1.18$ & $0.75$ & $0.71$ & $0.68$ & $1.54$ & $0.82$ & $0.63$ & $\textbf{0.62}$\\ 
 & $1.25$ & $0.80$ & $0.74$ & $0.70$ & $1.55$ & $0.92$ & $\textbf{0.65}$ & $0.67$\\ 
 & $1.66$ & $0.88$ & $0.70$ & $0.69$ & $2.15$ & $1.04$ & $\textbf{0.63}$ & $0.64$\\ 
 & $4.22$ & $1.55$ & $0.80$ & $0.70$ & $5.50$ & $2.15$ & $0.66$ & $\textbf{0.63}$\\
\hline 
\end{tabular}
}}
\end{table}

\subsubsection{Execution times comparison} 
We compared the execution times for both RANSAC and \RANSAAC together with their extra steps. By performing 100000 iterations on a dataset composed of 1000 inliers and 500 outliers with measurement noise of $\sigma\!=\!2$, we computed the total running time of each approach. Execution times are indicative and were measured on a 3.1GHz Intel Core i7 5557U processor. 
As seen from Table \ref{tab:RANSAACTimes}, the standard \RANSAAC method took $1\%$ more time than the original RANSAC. Moreover, if the LO-\RANSAAC variant is used, since only the hypotheses coming from the LO step are aggregated, the difference with respect to LO variant of RANSAC is almost negligible. 
\begin{table}[ht!]
\centering
\setlength{\tabcolsep}{3pt}
\caption{\small{Execution times for RANSAC, the last step minimization (LSM) using DLT, \RANSAAC and both the $wmean$ and $wgmed$ aggregation procedures using 100000 iterations, 1000/1500 inlier ratio and $\sigma\!=\!2$. Results are averages of 100 realizations.}}
{\small{
\begin{tabular}{c|c|c|c|c}
\hline 
RANSAC & LSM & \RANSAAC & $wmean$ agg. & $wgmed$ agg.\\ \hline 
52.810s & 0.007s & 53.269s & 0.007s & 0.052s \\ \hline 
\end{tabular}
}}
\label{tab:RANSAACTimes}
\end{table}

\subsubsection{Results by using adaptive termination}
\label{subsec:adaptiveTerminationResults}
We compared the LO-\RANSAAC algorithm using both proposed aggregation schemes together with USAC, RANSAC and RANSAC+M. A maximum of 1000 iterations was allowed for all methods, although the methods with adaptive termination varied this parameter dynamically during the execution, based on Eq.~\eqref{eq:amountOfIterations} using $\eta_0\!=\!0.99$. Results show average errors for 100 executions, where the noise value was fixed to $\sigma\!=\!5$ and the distance values were set for every method according to section \ref{sec:distanceParam}. USAC results are duplicated since adaptive termination was always activated.

\begin{figure}
\centering
	\includegraphics[width=.43\textwidth, height=4.2cm, clip, trim={0 0 10 5}]{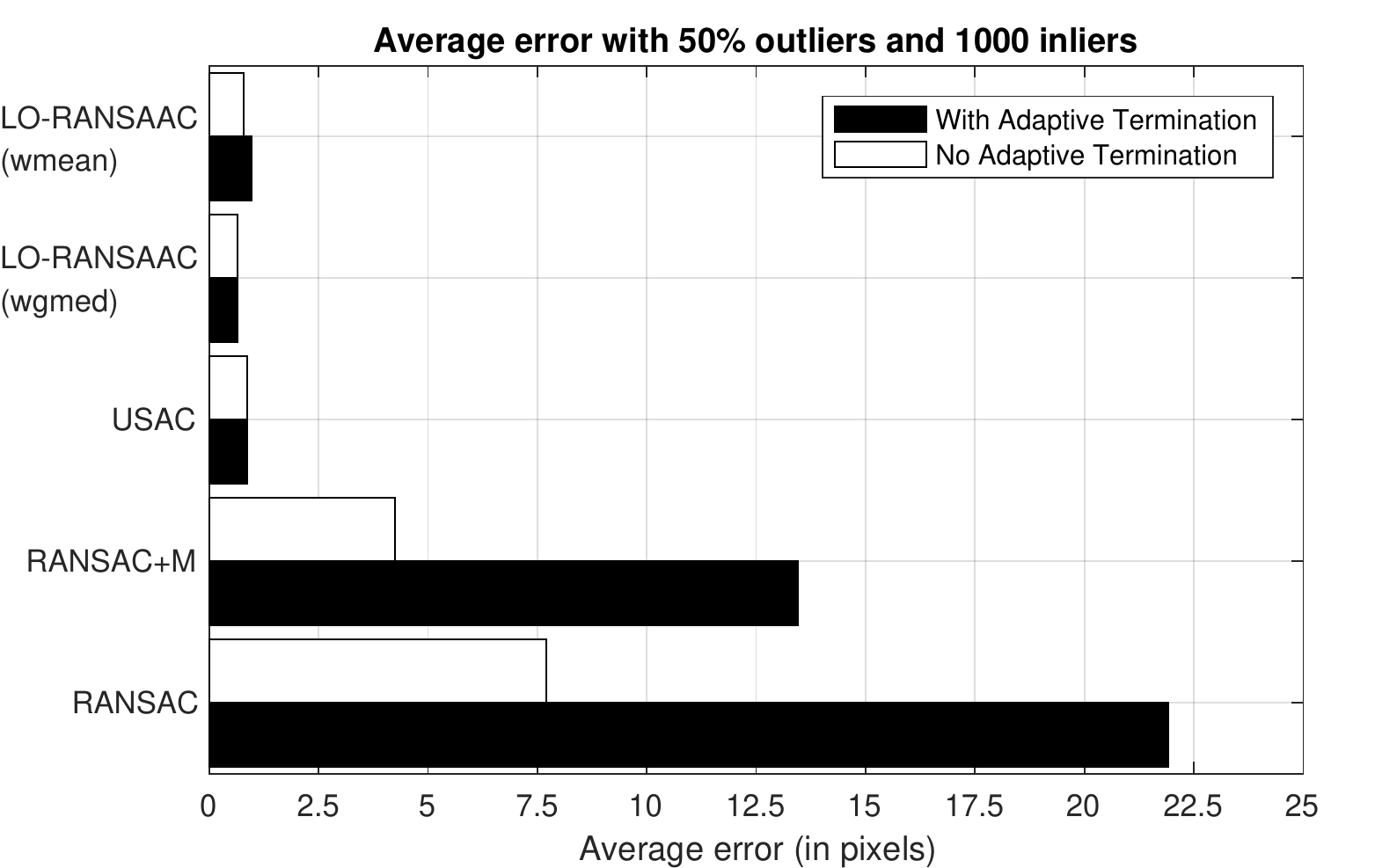}
\caption{Comparison by using adaptive termination to the proposed approach.}
\label{fig:adaptiveTermination}
\end{figure}

Several conclusions are drawn from results shown in Fig.~\ref{fig:adaptiveTermination}. First, while the weighted mean aggregation slightly suffers from using adaptive termination, the weighted geometric median aggregation in fact slightly benefits from it. The reason for this is again the same as mentioned before, and is related with the fact that making more iterations implies also having to better discriminate between samples. 

\subsubsection{Comparison with state-of-the-art methods.}
\label{subsec:comparisonWithStateOfTheArt}
We compared the proposed LO-\RANSAAC approach with RANSAC and several state-of-the-art variants, namely LORANSAC, MSAC \cite{torr_mlesac_2000}, ZHANGSAC \cite{Zhang_95} and the recent USAC algorithm \cite{raguram_usac_2013}. We included in the comparison two minimization methods applied directly on the inliers using an oracle: the linear least squares methods using the DLT algorithm and the non-linear method that minimizes the Sampson's approximation to the geometric reprojection error \cite{Hartley_2003}. For every method, we also evaluated adding the final minimization step (using the DLT algorithm) and kept only the best between both. 

\begin{figure*}
\centering
\hspace{-2.5mm}
\captionsetup[subfigure]{labelformat=empty}
\begin{subfigure}{0.5\textwidth}
  \centering
  \includegraphics[width=.9\linewidth,clip, trim={20 0 30 0}]
  {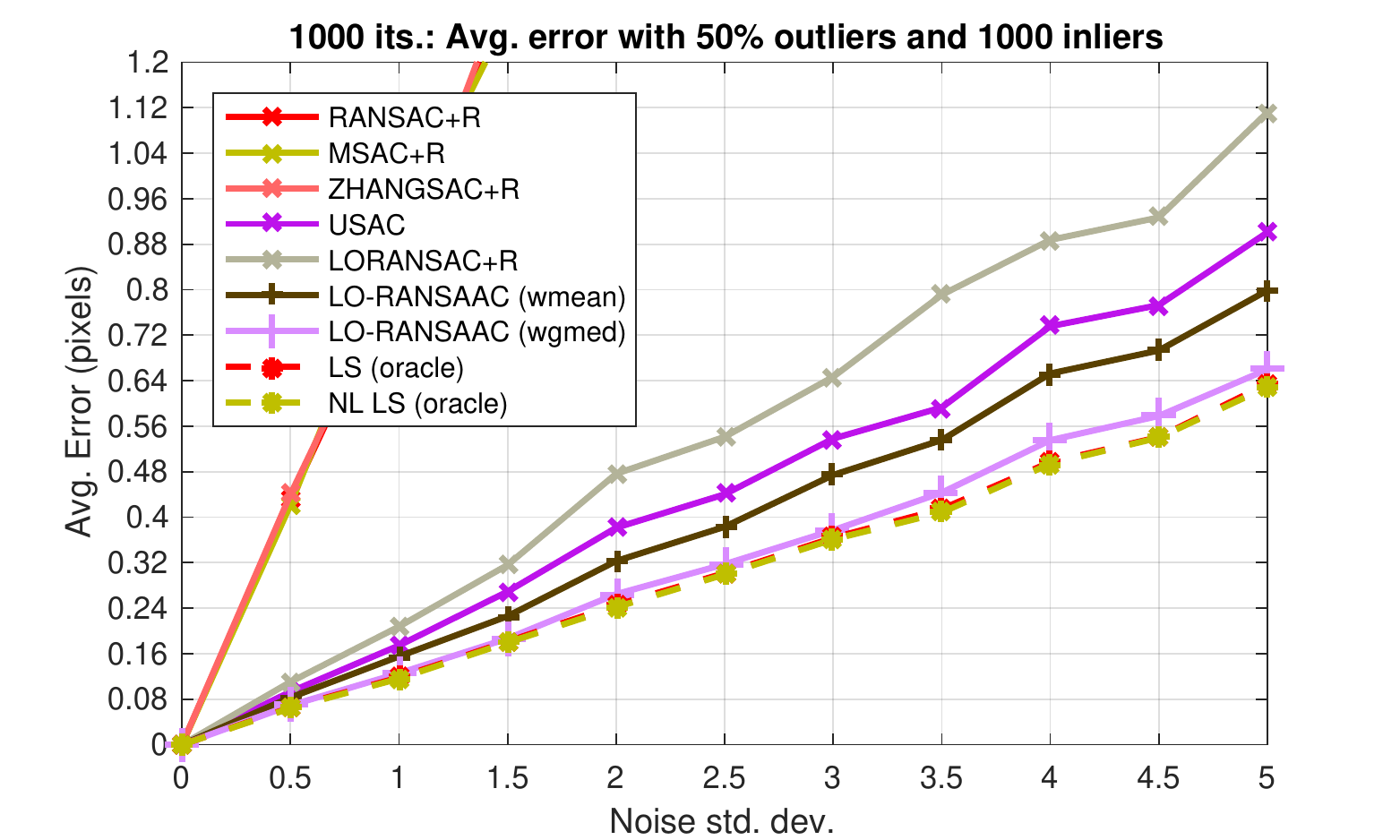}
  \caption{\small{$\bar{E}$: 1000 its with 1000/2000 inlier ratio.}}
\end{subfigure}%
\begin{subfigure}{0.5\textwidth}
  \centering
  \includegraphics[width=.9\textwidth, clip, trim={5 0 40 0}]
  {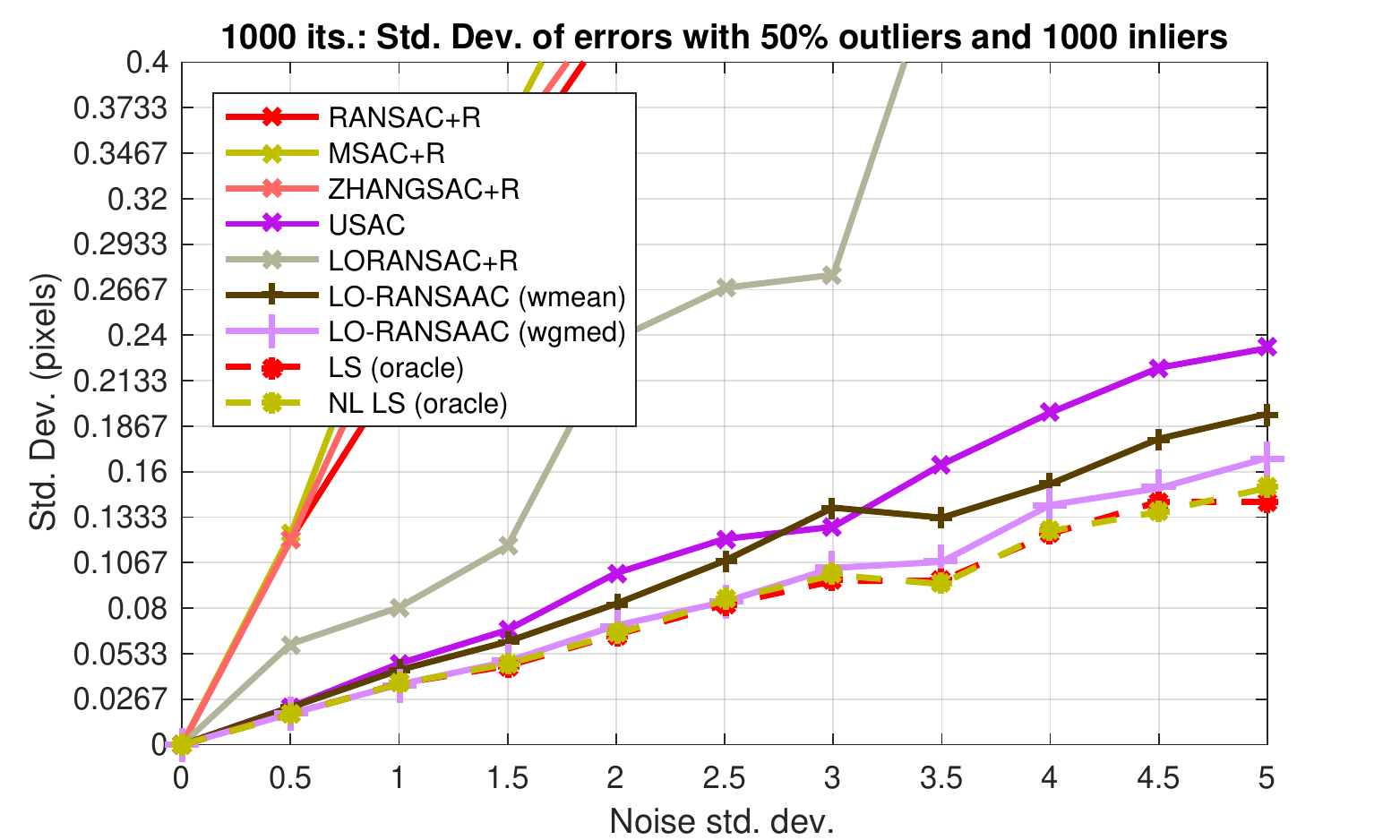}
  \caption{\small{$\sigma_E$: 1000 its with 1000/2000 inlier ratio.}}
\end{subfigure}
\begin{subfigure}{0.5\textwidth}
  \centering
  \includegraphics[width=.9\linewidth, clip, trim={20 0 30 0}]
  {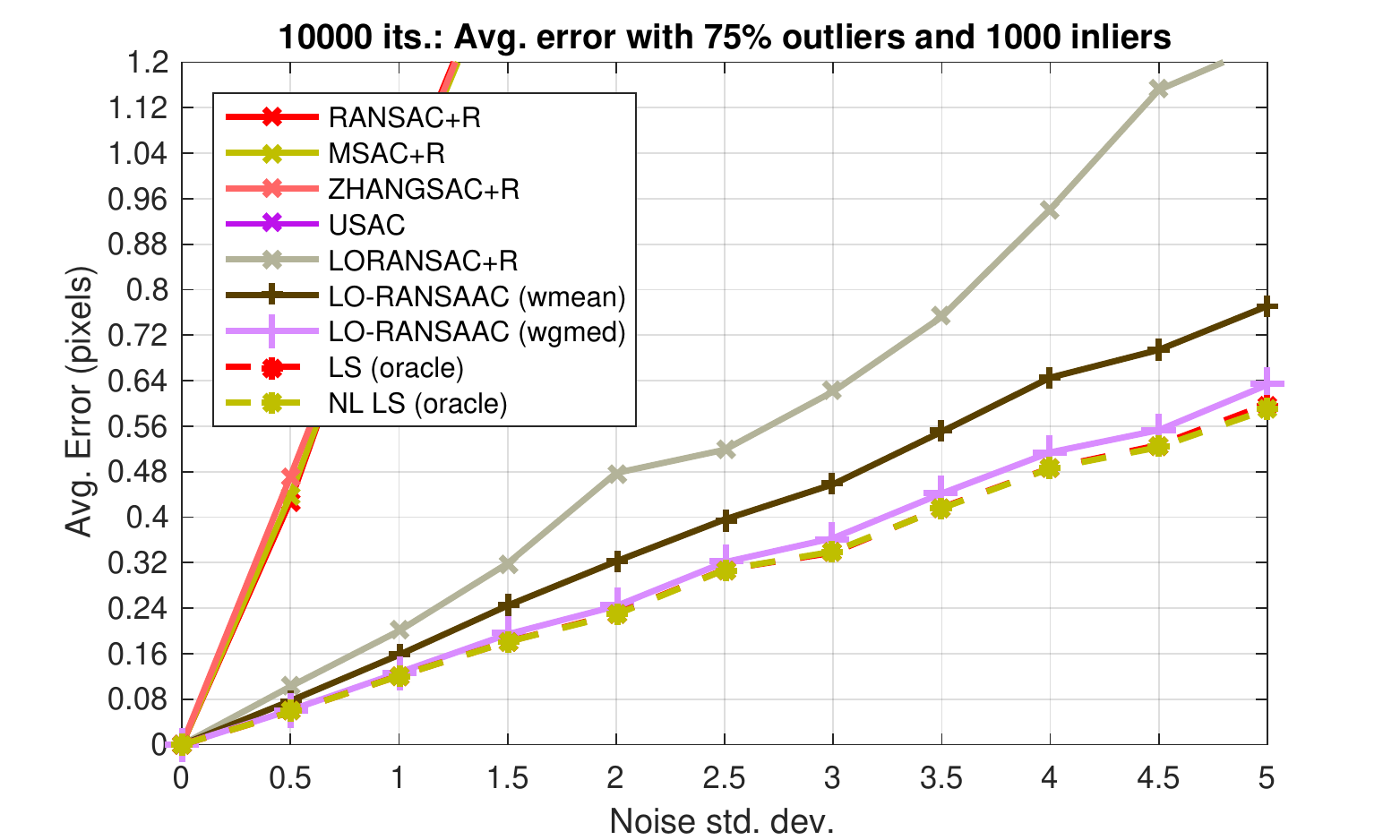}
  \caption{\small{$\bar{E}$: 10000 its with 1000/4000 inlier ratio.}}
\end{subfigure}%
\begin{subfigure}{0.5\textwidth}
  \centering
  \includegraphics[width=.9\textwidth, clip, trim={5 0 40 0}]
  {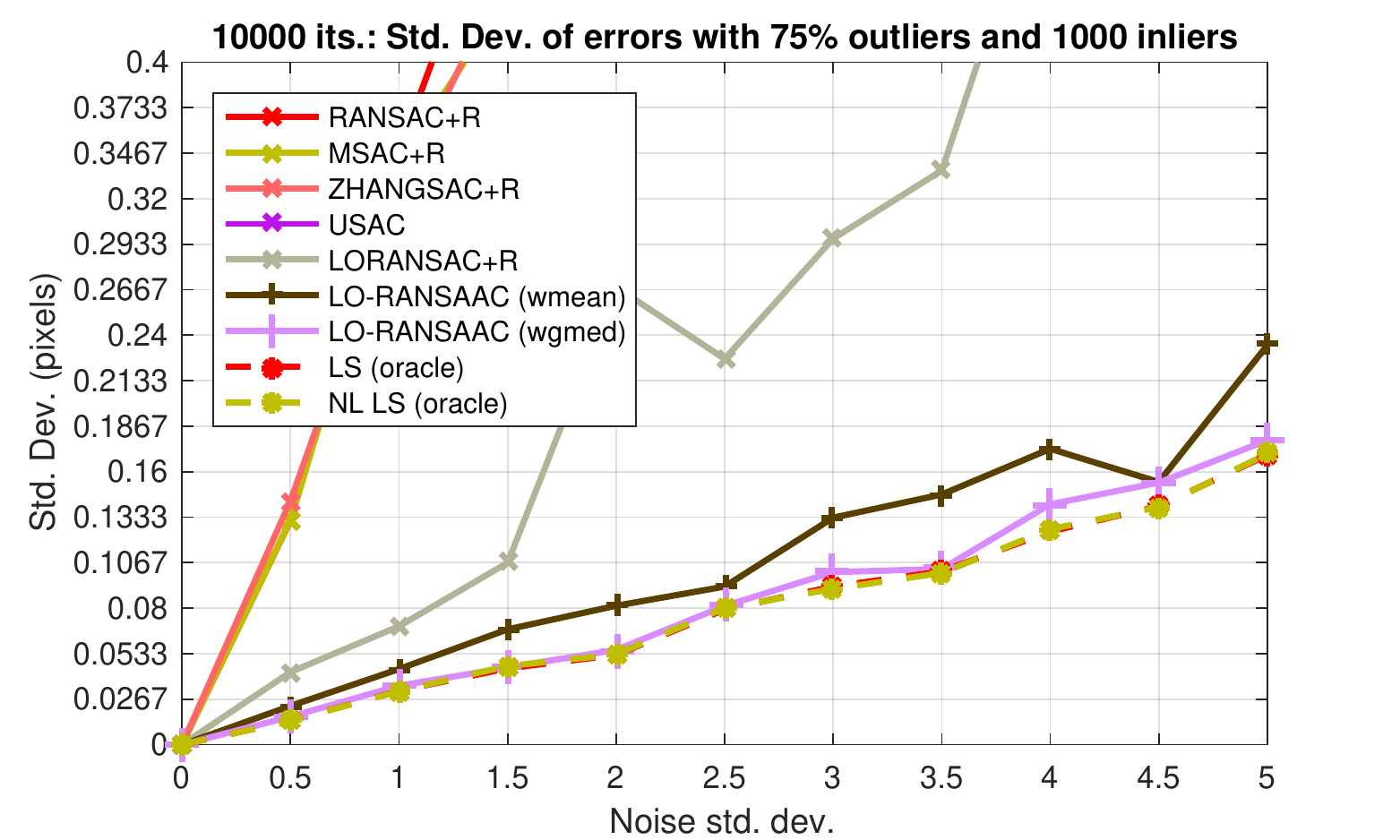}
  \caption{\small{$\sigma_E$: 10000 its with 1000/4000 inlier ratio.}}
\end{subfigure}
  \caption{\small{Avg. errors $\bar{E}$ and their std. dev. $\sigma_E$ by varying noise for several RANSAC and \RANSAAC variants. Experiment: 1000 inliers, different amounts of outliers and doing 1000/10000 iterations. 
  \textbf{First row}: 50\% outliers. \textbf{Last row}: 75\% outliers and 10k iterations.}}
  \label{fig:compareRANSAACWithAll1000Reg}
\end{figure*}

With 1000 inlier matches available, the average errors together with their standard deviations are shown in Fig.~\ref{fig:compareRANSAACWithAll1000Reg} for 50\% and 75\% outlier ratios. 
All approaches not using local optimization steps have much higher errors proving the importance of this method. The LO-RANSAC approach was able to attain good results, although it never achieves the accuracy of the USAC method. Every \RANSAAC variant improved over USAC, and strikingly, the $wgmed$ aggregation variant is close in performance to the methods that use an oracle. What is more, when there are 75\% outliers, the USAC approach failed while the LO-\RANSAAC method still achieves highly accurate and stable results.

\subsubsection{Performance under high outlier ratios}
\label{subsec:performanceHighOutlierRANSAAC}
We tested our method under extreme conditions with high outlier ratios. We evaluated the accuracy of different algorithms with 1000 inliers and 9000 outliers, using both 10000 and 20000 iterations. In table \ref{tab:highoutlierRatio} results are shown under two noise levels: mild/moderate noise ($\sigma\!=\!2$) or high noise ($\sigma\!=\!5$). Even under these extreme conditions, the LO-\RANSAAC method was still able to achieve extremely low error values. On the contrary, the RANSAC method failed considerably, even after applying the last step minimization, while LO-RANSAC did not achieve errors of under a pixel in average even by using 20000 iterations. Furthermore, USAC did not return any results. Interestingly, the achieved precision of the LO-\RANSAAC is close to the accuracy obtained by using an oracle. 

\newcolumntype{B}{>{\centering\arraybackslash}p{4em}}
\newcolumntype{C}{>{\centering\arraybackslash}p{3em}}
\newcolumntype{D}{>{\centering\arraybackslash}p{2em}}
\newcolumntype{E}{>{\centering\arraybackslash}p{2.2em}}

\begin{table}
\centering
\setlength{\tabcolsep}{2pt}
\caption{\small{High outlier ratio test: avg. errors for both aggregation methods of LO-\RANSAAC, compared with RANSAC+M, LO-RANSAC, USAC and computing LS on the inliers with an oracle. For noises $\sigma\!=\!2$ and $5$, each method was evaluated with 1000 inliers using both 10000 and 20000 iterations and 90\% outliers. Averaged over 50 realizations.}}
{\small{
%
\begin{tabular}{|c|c|c|c|c|c|c|c|c|c|c|c|}
\hline 
$\sigma$ &  \multicolumn{2}{ c| }{$wmean$} & \multicolumn{2}{ c| }{$wgmed$} & \multicolumn{2}{ c| }{\scriptsize{RANSAC+M}} & \multicolumn{2}{ c| }{\scriptsize{LO-RANSAC}} & \multicolumn{2}{ c| }{\scriptsize{USAC}} & {\scriptsize{Oracle}}\\ \hline 
 & 10k & 20k  & 10k & 20k  & 10k & 20k  & 10k & 20k  & 10k & 20k  & - \\ \hline 
\multirow{1}{*}{2}  & $0.53$ & $0.46$ & $\textbf{0.36}$ & $\textbf{0.31}$ & $49.15$ & $19.90$ & $16.68$ & $1.76$ & $-$ & $-$ & $\textbf{0.28}$\\ 
\hline 
\multirow{1}{*}{5} & $1.35$ & $1.38$ & $\textbf{0.94}$ & $\textbf{1.15}$ & $23.94$ & $27.69$ & $4.35$ & $6.30$ & $-$ & $-$ & $\textbf{0.74}$\\ 
\hline 

\end{tabular}
}}
\label{tab:highoutlierRatio}
\end{table}

\subsubsection{Performance evaluation using random homographies}
\label{subsec:performanceRandomHomographies}
We also tested the robustness of the proposed approach to different homographies. We used the Zuliani toolbox available online \cite{Zuliani08a} to simulate random homographies. 

\begin{figure*}
\vspace{-2mm}
\centering
\captionsetup[subfigure]{labelformat=empty}
\begin{subfigure}{0.5\textwidth}
  \centering
	\includegraphics[width=.9\textwidth, clip, trim={10 0 30 0}]
	{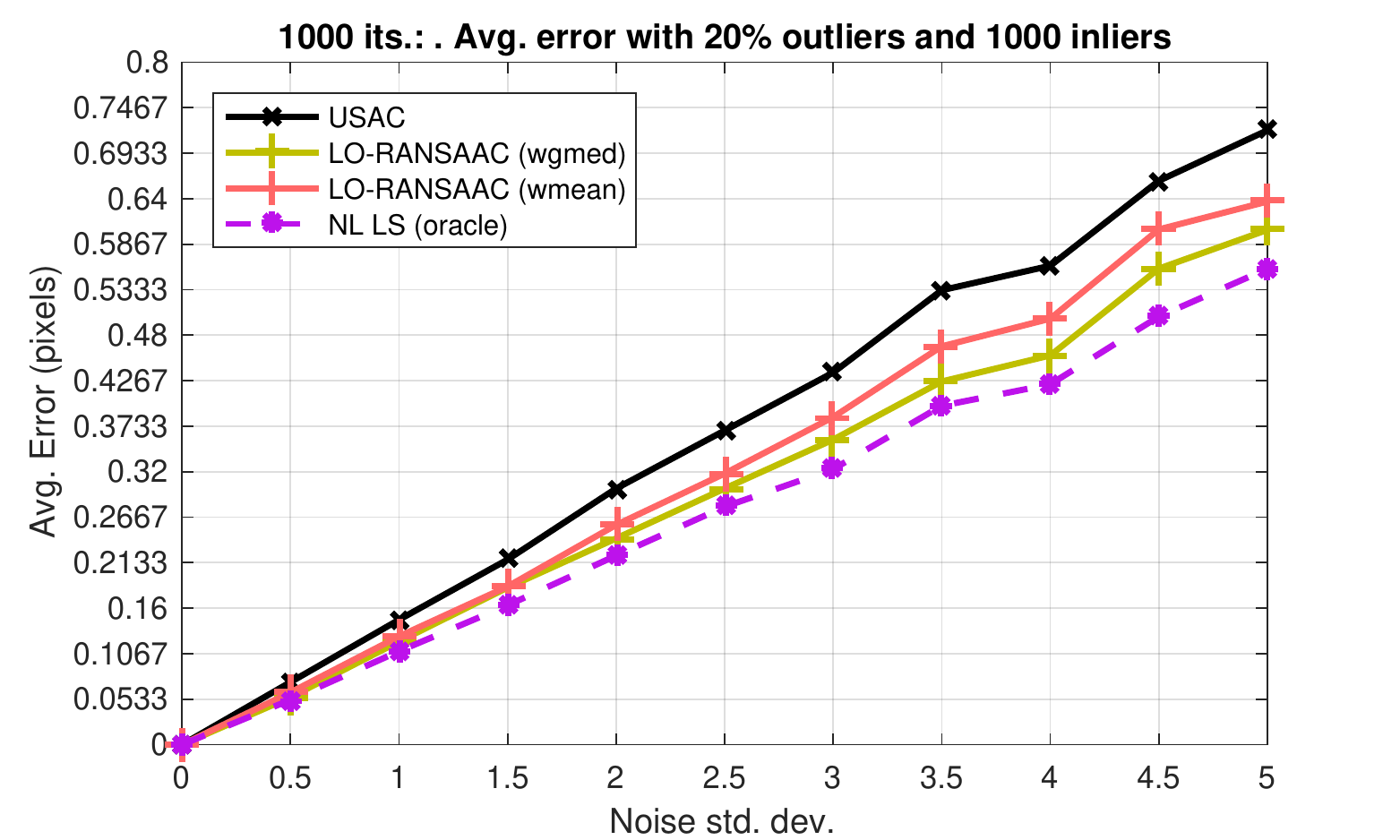}
  \caption{\small{$\bar{E}$: 1000 its with 1000/1250 inlier ratio.}}
\end{subfigure}%
\begin{subfigure}{0.5\textwidth}
  \centering
  \includegraphics[width=.9\linewidth, clip, trim={5 0 40 0}]
  {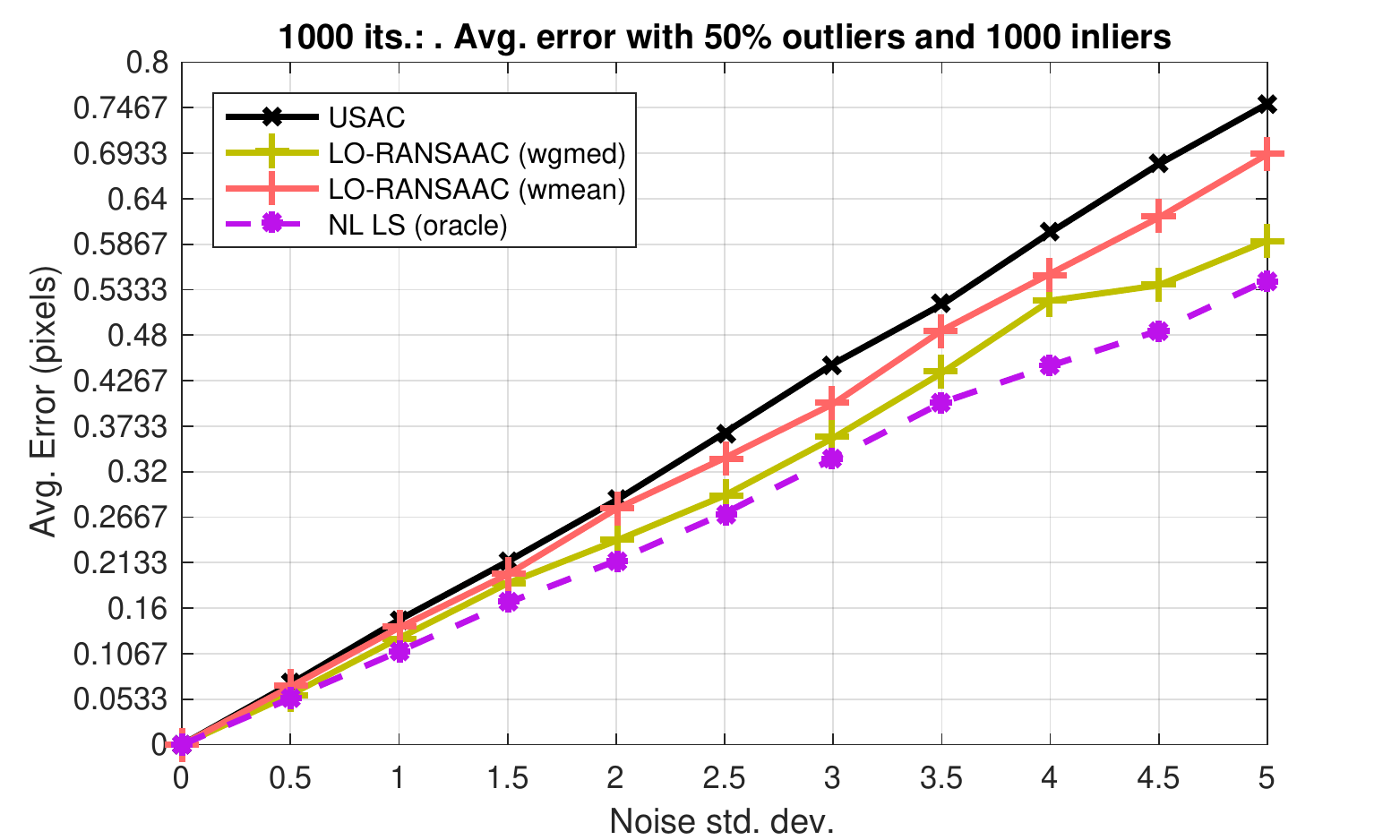}
  \caption{\small{$\bar{E}$: 1000 its with 1000/2000 inlier ratio.}}
\end{subfigure}%

  \caption{\small{Avg. errors $\bar{E}$ by varying the homography and the input matches for each trial, using 1000 iterations. \textbf{Left}: 20\% outliers and 1000 inliers. \textbf{Right}: 50\% outliers and 1000 inliers.}}
  \label{fig:resRandomHomo}
\end{figure*}

Then, we evaluated both LO-\RANSAAC aggregation methods together with USAC \cite{raguram_usac_2013} which turned out to be the best approach during our previous tests and again, aided by an oracle, the least squares minimization of the Sampson approximation to the geometric reprojection error. The experiment again consisted in averaging the error over 100 trials by varying the noise and the inlier/outlier configuration. The amount of iterations was set to 1000. Differing from the previous experiments where both the homography and the inliers were fixed, a new random homography and random inliers sampled from it were generated for each trial. Then, as before, outliers were injected in the input data by picking random positions on both images, and finally Gaussian white noise was added to the final points. Results confirmed the improvement of LO-\RANSAAC over USAC as shown in Fig.~\ref{fig:resRandomHomo}, validating the versatility of the approach. 

As a final test, instead of fixing the amount of iterations performed, we evaluated the same methods using adaptive termination (see sec. \ref{sec:adaptiveTermination}). Again the maximum number of iterations was set to 1000, however in practice the algorithm performed around 12 and 85 iterations for 20\% and 50\% outliers respectively, together with a single LO step. Interestingly, even by taking such low amount of iterations, the method still improved over USAC. However, this did not occur using $wmean$ aggregation for the case of 1000 inliers, due to its already observed necessity to get more models in order for the average to be closer to the optimal solution. A possible workaround for this, validated empirically, was to double the theoretical number of iterations given by Eq.~\eqref{eq:amountOfIterations} in the adaptive procedure. On the contrary, the LO-\RANSAAC method with $wgmed$ aggregation obtained more accurate results than USAC on every case. 

Lastly, we observed some instability for our approach when using 100/200 inlier ratio and $\sigma\!=\!5$. This peak was caused by some homographies which were not correctly averaged. In fact, some degenerate cases of homographies would not be correctly aggregated using the approach described in section \ref{sec:aggregation}. 
Indeed, averaging points close to the horizon of the homography (i.e., the line of points that are projected to the infinity) will be unstable. 
Because of the noise on the input data, each all-inlier homography will be noisy. Then it is possible for one of the preselected points to lie close but on different sides of the horizon line for two different valid homographies, so when projected, they end up on opposite sides of the image and their average does not follow the geodesics of the space. 
To correct this behaviour, it would suffice to early detect which preselected points are close to the horizon line and either avoid averaging them or restart the algorithm by preselecting new points away from the horizon.
Nevertheless, for several applications such as panorama generation, the probability of this situation to occur is very low, since images are taken continously.

\begin{figure*}
\centering
\captionsetup[subfigure]{labelformat=empty}
\begin{subfigure}{0.5\textwidth}
  \centering
  \includegraphics[width=.9\linewidth, clip, trim={20 0 30 0}]{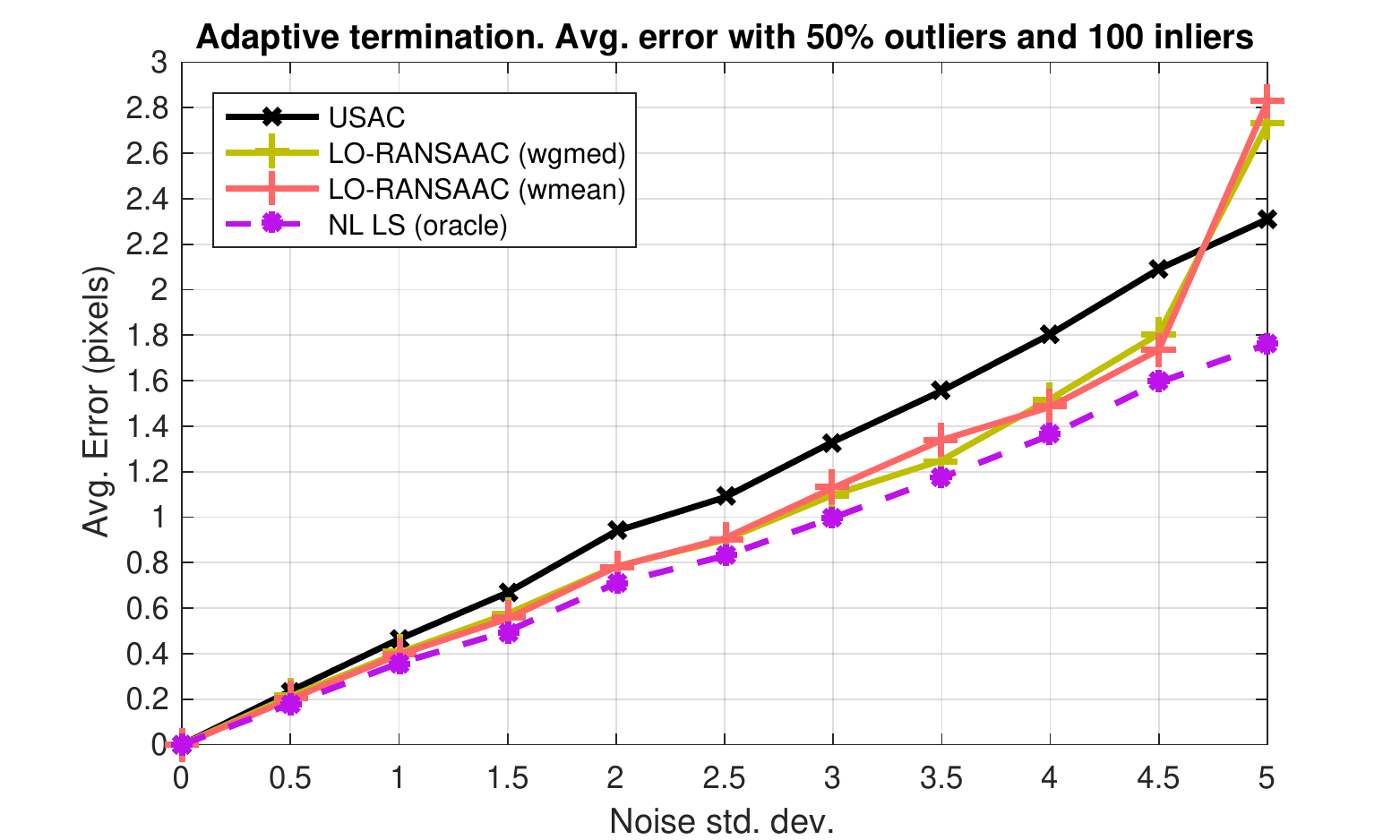}
  \caption{\small{$\bar{E}$: 100/200 inlier ratio.}}
\end{subfigure}%
\begin{subfigure}{0.5\textwidth}
  \centering
  \includegraphics[width=.9\linewidth, clip, trim={10 0 40 0}]{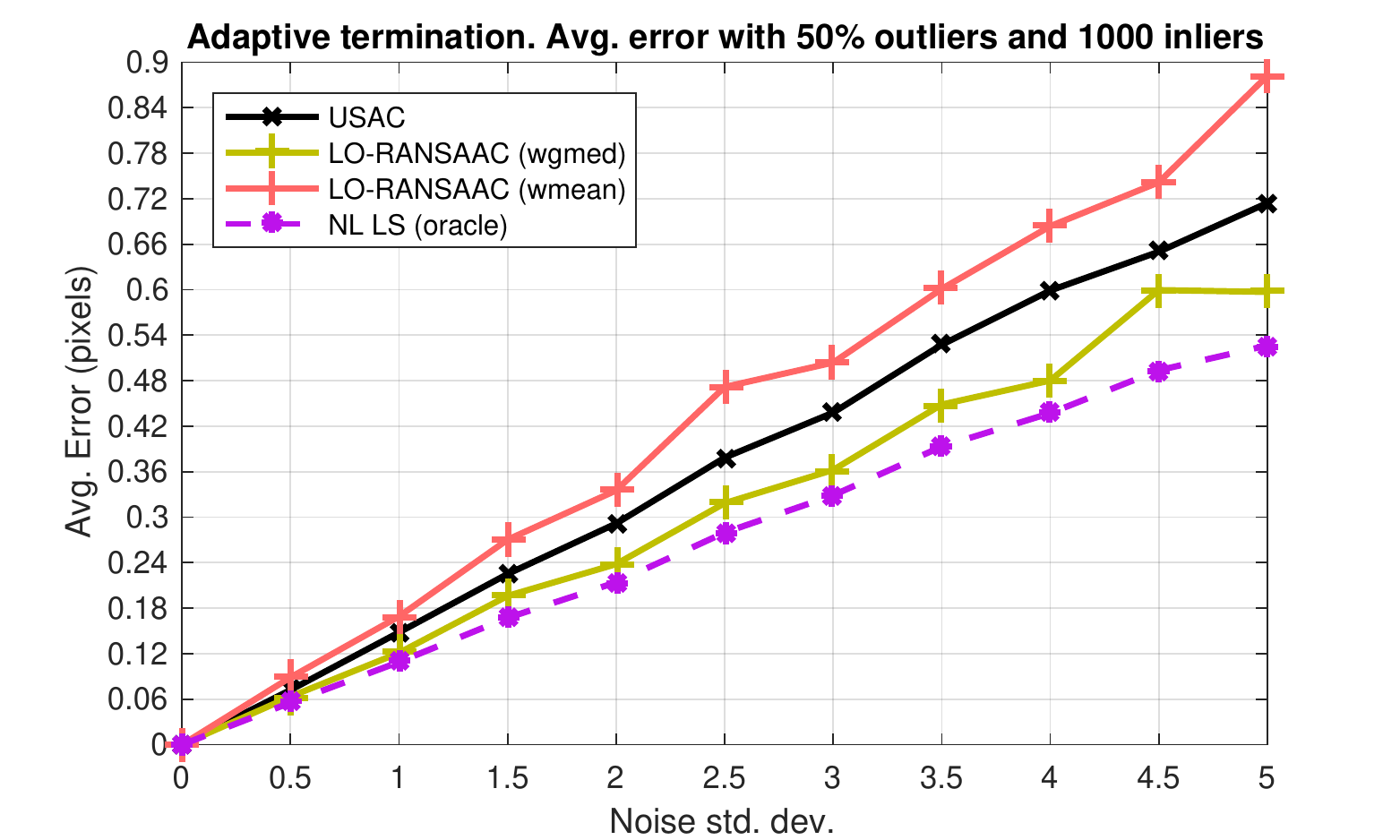}
  \caption{\small{$\bar{E}$: 1000/2000 inlier ratio.}}
\end{subfigure}

  \caption{\small{Avg. errors $\bar{E}$ by varying the homography and the input matches for each trial, using adaptive termination and 50\% outliers. \textbf{Left}: 100 inliers. \textbf{Right}: 1000 inliers.}}
  \label{fig:resRandomHomoAdaptive}
\end{figure*}

\subsection{Application 2: Estimating Homography+Distortion}
\label{sec:HomoPlusDistortion}
The recently proposed method \cite{Kukelova_2015} to estimate an homography and a distortion coefficient for each image was evaluated to test the robustness of the proposed approach against different model noises. In this case, the available algorithm to compute such model assumes only 5 input matches, and it is not possible to perform a least squares like minimization based on the authors supplied code. Therefore, this is an interesting case in which, evidently, no local optimization is possible as is, and no last step minimization could be applied. Therefore, we restricted ourselves to evaluating the original \RANSAAC approach using $wmean$ aggregation, comparing it with RANSAC, as in the author's work.

\begin{figure*}
\centering
\captionsetup[subfigure]{labelformat=empty}
\begin{subfigure}{0.50\textwidth}
  \centering
  \includegraphics[width=.85\linewidth, height=4.8cm,trim={3 3 30 0}, clip]{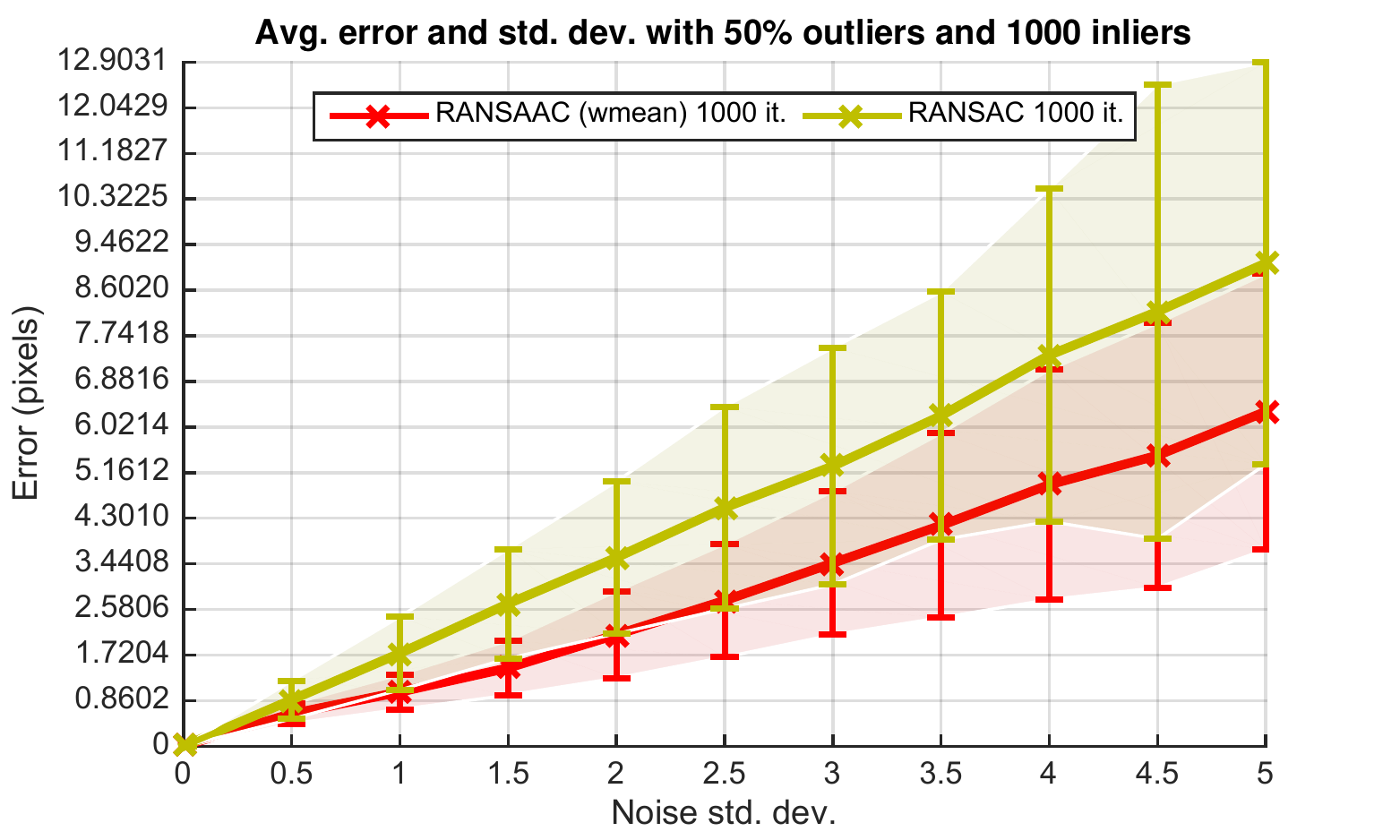}
  \caption{\small{1000 its - 1000/2000 inl. ratio}}
\end{subfigure}%
\begin{subfigure}{0.50\textwidth}
  \centering
  \includegraphics[width=.85\linewidth, height=4.8cm, trim={8 3 30 0}, clip]{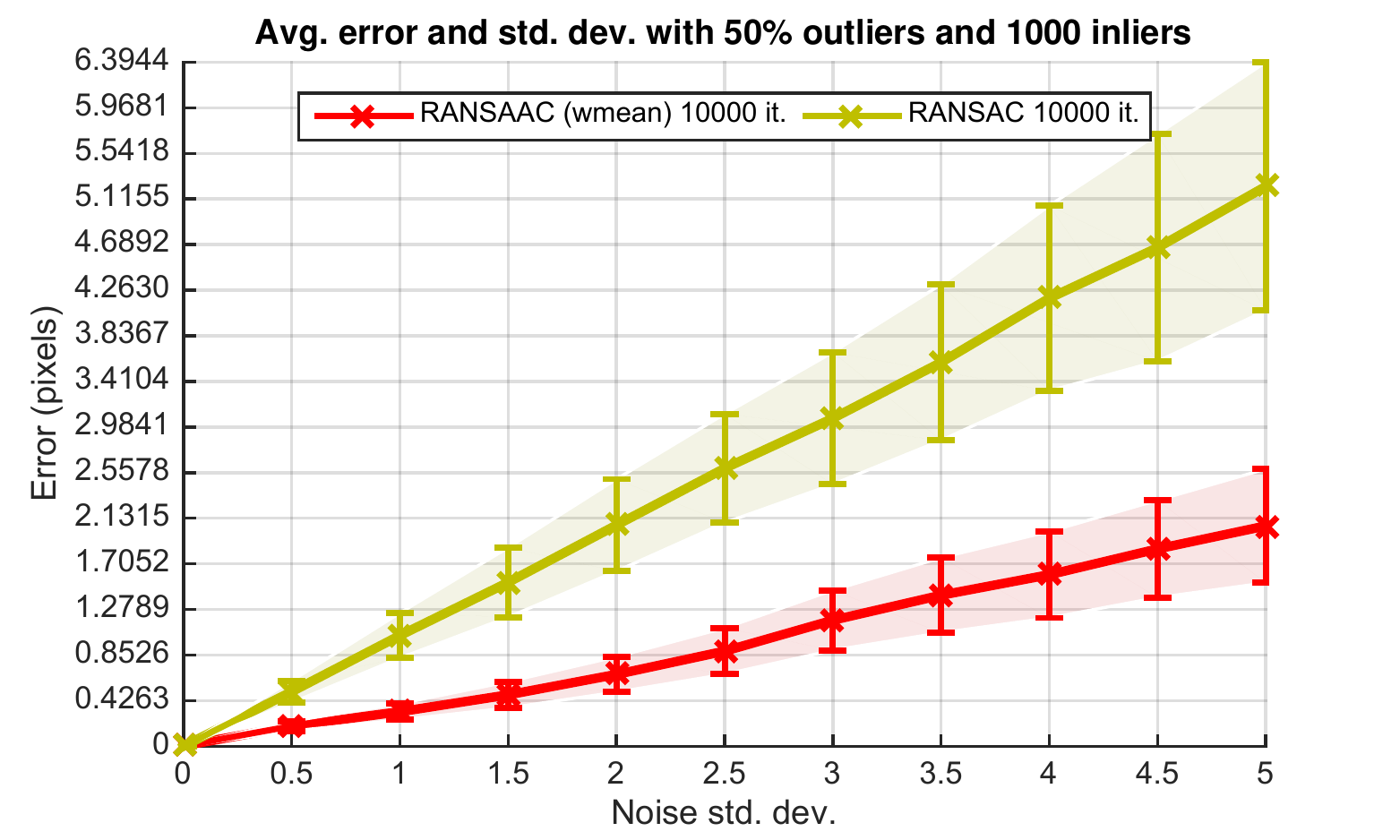}
  \caption{\small{10000 its - 1000/2000 inl. ratio}}
\end{subfigure}
\caption{\small{Avg. errors $\bar{E}$ and their std. dev. $\sigma_E$ by varying noise for RANSAC and \RANSAAC (weighted mean) with 50\% outliers, different iterations and quantity of inliers for the $H_5$ transformation \cite{Kukelova_2015}.}}
  \label{fig:compareRANSACandWMEANH5}
\end{figure*}

As seen in Fig.~\ref{fig:compareRANSACandWMEANH5}, the method outperforms RANSAC in every evaluated experiment on both accuracy and stability, and the difference in performance gets higher as the amount of iterations increase, particularly for higher outlier percentages. This experiment proves the versatility of our approach compared to other more complex models. 

\section{Conclusion}
\label{sec:conclusions}
In this article we introduced a simple, yet powerful RANSAC modification that improves the method 
by combining the random consensus idea using samples with minimal cardinality with a statistical approach performing an aggregation of estimates. This comes with an almost negligible extra computational cost. The most interesting advantage of our approach is that it makes better use of the generated hypotheses, obtaining improved results when using the exact same hypotheses.
What is more, most of RANSAC enhancements easily fit into the proposed method. By adding local optimization to \RANSAAC (and using weighted geometric median  aggregation), the resulting accuracy and stability increased considerably surpassing every other RANSAC variant. Moreover, it succeeded in accurately estimating models under 90\% outliers, a situation where most 
 approaches failed. 

As a future work, 
we plan to extend the method to the case of epipolar geometry estimation. 

\ifCLASSOPTIONcompsoc
  \section*{Acknowledgments}
\else
  \section*{Acknowledgment}
\fi
Work partly founded by the Centre National d'Etudes Spatiales (CNES, MISS Project), the Office of Naval research (ONR grant N00014-14-1-0023) and by the Ministerio de Ciencia e Innovaci\'{o}n under grant TIN2014-53772-R. During this work, Martin Rais had a fellowship of the Ministerio de Economia y Competividad (Spain), reference BES-2012-057113, for the realization of his Ph.D. thesis.

\ifCLASSOPTIONcaptionsoff
  \newpage
\fi




\bibliographystyle{IEEEtran}
\bibliography{references}

%
%

%

\vspace{-4mm}
\begin{IEEEbiography}[{\includegraphics[width=1in,height=1.2in,clip,keepaspectratio]{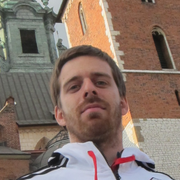}}]{Martin Rais}
received his B.Sc. from Universidad de Buenos Aires, Argentina, and his M.Sc. from the Ecole Normale Sup\'{e}rieure de Cachan, France in 2012. He is currently fulfilling a joint Ph.D. between the DMI, University of the Balearic Islands, Spain and the CMLA, ENS Paris Saclay, France. His research interests include image registration, denoising, inpainting, stereovision, remote sensing and machine learning. 
\end{IEEEbiography}
\vspace{-8mm}
\begin{IEEEbiographynophoto}
{Gabriele Facciolo}
received his B.Sc. and M.Sc. from Universidad de la Republica del Uruguay, and his Ph.D. (2011) from Universitat Pompeu Fabra, Spain. He joined CMLA at the Ecole Normale Sup\'{e}rieure de Cachan in 2011 where he is currently associate research professor. His main areas of research interest are image denoising, inpainting, stereovision, and remote sensing. 
\end{IEEEbiographynophoto}
\vspace{-8mm}
\begin{IEEEbiographynophoto}{Enric Meinhardt-Llopis}
Enric Meinhardt-Llopis received his B.Sc. in mathematics from the Technical University of Catalonia (2003, Barcelona), his M.Sc from \'{E}cole Normale Sup\'{e}rieure de Cachan (2006, Cachan, France) and his Ph.D from Universitat Pompeu Fabra (2011, Barcelona). After a post-doc in the group of Jean-Michel Morel, he became associate professor at the ENS Cachan in 2014.
\end{IEEEbiographynophoto}
\vspace{-8mm}
\begin{IEEEbiographynophoto}
{Jean-Michel Morel}
received his PhD degree in applied mathematics from University Pierre et Marie Curie, Paris, France in 1980. He started his career in 1979 as assistant professor in Marseille Luminy, then moved in 1984 to University Paris Dauphine where he was promoted professor in 1992. He is Professor of Applied Mathematics at the Ecole Normale Sup\'{e}rieure de Cachan since 1997. His research is focused on the mathematical analysis of image processing. He is a 2015 laureate of the Longuet-Higgins prize and of the CNRS {\it m\'{e}daille de l'innovation}.
\end{IEEEbiographynophoto}
\vspace{-8mm}
\begin{IEEEbiographynophoto}
{Antoni Buades}  received the Ph.D. degree in applied mathematics from the Universitat de les Illes Balears, Spain, in 2006. He is currently a Researcher with Universitat Illes Balears and ENS Cachan. His research is focused on the mathematical analysis of image processing.
\end{IEEEbiographynophoto}
\vspace{-8mm}
\begin{IEEEbiographynophoto}
{Bartomeu Coll}
received the Ph.D.  degree in applied mathematics from the Autonoma University of Barceloba, Spain, in 1987. In 1989, he received a potsdoctoral fellowship to begin research in the field of digital image processing at the laboratory of Ceremade, Paris, France. He is currently a full professor with Universitat Illes Balears. His research is focused on the study of mathematical models for image processing and applications.
\end{IEEEbiographynophoto}

\vfill




\end{document}